\documentclass[lettersize,journal]{IEEEtran}
\usepackage{amsmath,amsfonts}
\usepackage{algorithmic}
\usepackage{algorithm}
\usepackage{array}
\usepackage[caption=false,font=normalsize,labelfont=sf,textfont=sf]{subfig}
\usepackage{textcomp}
\usepackage{stfloats}
\usepackage{url}
\usepackage{verbatim}
\usepackage{graphicx}
\usepackage{cite}
\usepackage{times}
\usepackage{epsfig}
\usepackage{graphicx}
\usepackage{amsmath}
\usepackage{amssymb}
\usepackage{bm}
\usepackage{nicefrac} 
\usepackage{microtype} 
\usepackage{blindtext}
\usepackage{tabulary,multirow,overpic}
\usepackage{booktabs}
\usepackage{makecell}  % xline
\usepackage{color}

\begin{document}
%%%%%%%%%%%%%%%%%%%%%%%%%%%%%%%%%%%%%%%% title %%%%%%%%%%%%%%%%%%%%%%%%%%%%%%%%%%%%%%%%
\title{Detect Any Shadow: Segment Anything for \\ Video Shadow Detection}

\author{
	\thanks{This work was supported by NSFC under Contract U20A20183 and 62021001. 
		It was also supported by GPU cluster built by MCC Lab of Information Science and Technology Institution, USTC, and the Supercomputing Center of the USTC. (Corresponding authors: Wengang Zhou and Houqiang Li.)}
	
	Yonghui Wang,
	Wengang Zhou$^{\dagger}$,~\IEEEmembership{Senior Member,~IEEE},
	Yunyao Mao,
	and~Houqiang Li$^{\dagger}$,~\IEEEmembership{Fellow,~IEEE}
	\IEEEcompsocitemizethanks{\IEEEcompsocthanksitem Yonghui Wang, Wengang Zhou, Yunyao Mao, and Houqiang Li are with the CAS Key Laboratory of Technology in Geo-spatial Information Processing and Application System, Department of Electronic Engineering and Information Science, University of Science and Technology of China, Hefei, 230027, China.
	Wengang Zhou and Houqiang Li are also with Institute of Artificial Intelligence, Hefei Comprehensive National Science Center. 
	E-mail: \{wyh1998, myy2016\}@mail.ustc.edu.cn; \{zhwg, lihq\}@ustc.edu.cn.
	% \IEEEcompsocthanksitem  $^{\dagger}$The first two authors contribute equal to this work.
	\IEEEcompsocthanksitem  $^{\dagger}$Corresponding authors: Wengang Zhou and Houqiang Li.
}}

%\author{IEEE Publication Technology,~\IEEEmembership{Staff,~IEEE,}

%\thanks{This paper was produced by the IEEE Publication Technology Group. They are in Piscataway, NJ.}% <-this % stops a space
%\thanks{Manuscript received April 19, 2021; revised August 16, 2021.}}

\markboth{IEEE TRANSACTIONS ON CIRCUITS AND SYSTEMS FOR VIDEO TECHNOLOGY,~Vol.~**, No.~**, September~2023}%
{Shell \MakeLowercase{\textit{et al.}}: A Sample Article Using IEEEtran.cls for IEEE Journals}


\IEEEpubid{\begin{minipage}{\textwidth}\ \\[12pt] \centering Copyright~\copyright~2023 IEEE. Personal use of this material is permitted. 
However, permission to use this material for any other purposes must \\ be obtained from the IEEE by sending an email to pubs-permissions@ieee.org.
\end{minipage}}
\maketitle

%%%%%%%%%%%%%%%%%%%%%%%%%%%%%%%%%%%%%%%% abstract %%%%%%%%%%%%%%%%%%%%%%%%%%%%%%%%%%%%%%%%
\begin{abstract}
Segment anything model (SAM) has achieved great success in the field of natural image segmentation.
Nevertheless, SAM tends to consider shadows as background and therefore does not perform segmentation on them.
In this paper, we propose ShadowSAM, a simple yet effective framework for fine-tuning SAM to detect shadows.
Besides, by combining it with long short-term attention mechanism, we extend its capability for efficient video shadow detection.
Specifically, we first fine-tune SAM on ViSha training dataset by utilizing the bounding boxes obtained from the ground truth shadow mask.
Then during the inference stage, we simulate user interaction by providing bounding boxes to detect a specific frame (\emph{e.g.}, the first frame).
Subsequently, using the detected shadow mask as a prior, we employ a long short-term network to learn spatial correlations between distant frames and temporal consistency between adjacent frames, thereby achieving precise shadow information propagation across video frames.
Extensive experimental results demonstrate the effectiveness of our method, with notable margin over the state-of-the-art approaches in terms of MAE and IoU metrics. 
% that our method outperforms the state-of-the-art techniques, pushing  Mean Absolute Error (MAE) to 0.020 (31.0\% point improvement) and Intersection over Union (IoU) to 0.693 (8.3\% point improvement), respectively.
Moreover, our method exhibits accelerated inference speed compared to previous video shadow detection approaches, validating the effectiveness and efficiency of our method.
The source code is now publicly available at \url{https://github.com/harrytea/Detect-AnyShadow}.
\end{abstract}


\section{Introduction}
\IEEEPARstart{S}{hadow} is a common phenomenon in natural scenes.
The existence of shadow provides various hints for scene understanding, \emph{e.g.}, light direction~\cite{lalonde2009est,lalonde2012estimating,panagopoulos2009robust}, scene geometry~\cite{huang2011what,takahiro2009attached,junejo2008estimating,okabe2009attached,thien2003image}, and camera parameters~\cite{wu2010camera,junejo2008estimating}.
Detecting shadows is beneficial for many other computer vision tasks, such as object detection~\cite{cucchiara2003detecting,chien2002efficient}, image segmentation~\cite{ecins2014shadow}, and virtual reality scene generation~\cite{liu2020arshadowgan}.
Recently, single-image shadow detection has achieved great progress due to the development of deep learning algorithms~\cite{vicente2016large,hu2018direction,zhu2018bidirectional,zheng2019distraction,chen2020multi,zhu2021mitigating}, while for shadow detection over dynamic scenes, the exploration is less.

\begin{figure}[t]
	\centering
	\includegraphics[width=0.98\linewidth]{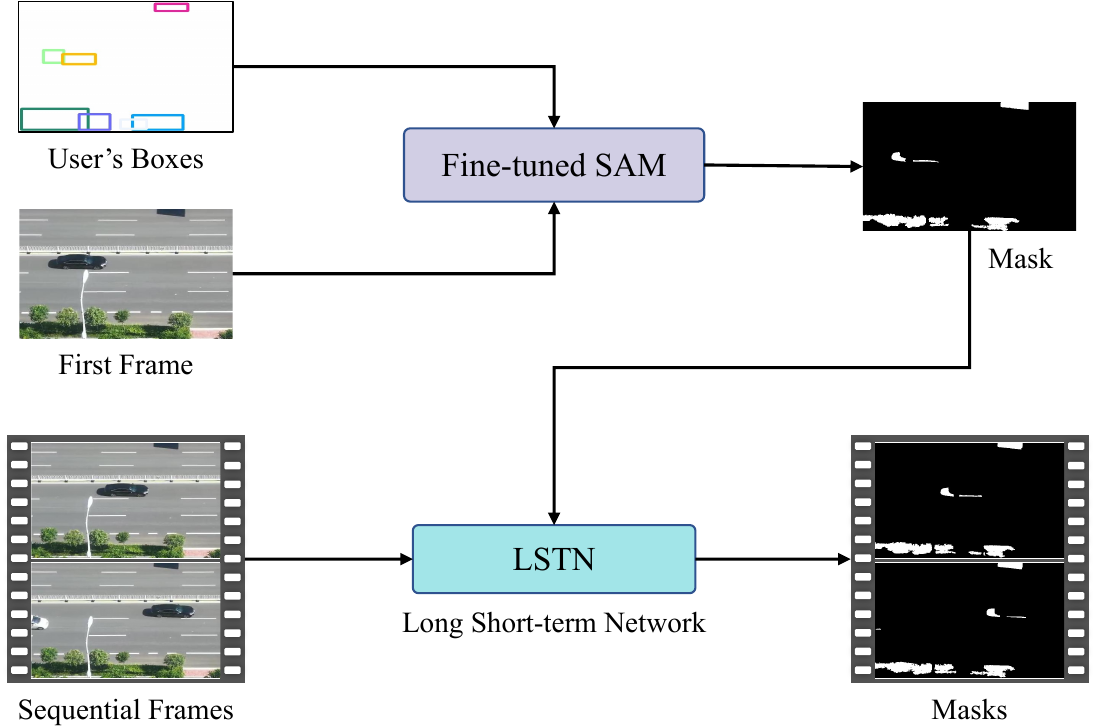}
	\caption{An overview of the proposed ShadowSAM. First, fine-tuned SAM takes the first frame of video and user's bounding boxes as inputs to predict the mask as shadow prior. Then, the long short-term network propagates this information across frames to facilitate the prediction of subsequent frames.}
	\label{fig:intro}
\end{figure}

Previous deep learning-based video shadow detection methods directly model the feature correlation between frames, \emph{e.g.}, TVSD~\cite{chen2021triple} learns at both intra-frame and inter-frame levels to model their correlation features, and SC-Cor~\cite{ding2022learning} learns the correspondence among nearby frames at inter-frame level.
These methods have achieved promising results.
Nevertheless, they do not incorporate shadow prior information.
As a similar pixel-level video prediction task, video object segmentation (VOS)~\cite{pont20172017,xu2018youtube,yang2021associating,cheng2022xmem} tracks and segments objects of interest throughout the entire video based on the object prior mask given in the first frame.
Motivated by this, we consider predicting the shadow mask of a specific frame (\emph{e.g.}, the first frame), which we call it~\emph{shadow prior}, within a video and propagating it to the entire video by learning the correspondence between frames.
We argue that the introduction of shadow prior provides the network with more information to improve video shadow detection.
Moreover, by incorporating the shadow prior, the interaction with adjacent frames can be reduced during the detection of the current frame, resulting in faster inference speed.
For instance, in the case of TVSD~\cite{chen2021triple}, detecting shadows in the current frame requires the empirical selection of the subsequent five frames to facilitate inter-frame interactions, which increases the inference time of the network and hampers the trade-off between accuracy and speed.
In contrast, our method requires interaction with the first frame and the previous frame.
Therefore, by leveraging shadow prior and propagating it across the video, we can enhance the precision of the prediction network while concurrently boosting its computational efficiency.
% through effective propagation of the shadow prior, we can enhance the precision of the prediction network while concurrently boosting its computational efficiency.

\IEEEpubidadjcol

To obtain the shadow prior, \emph{i.e.}, shadow mask, from a specific frame in the video, a straightforward approach is to directly utilize existing single-image shadow detectors~\cite{chen2020multi,zhu2021mitigating}.
However, due to the scarcity of training data, their generalization capability is constrained, leading to prediction errors that will accumulate during video propagation~\cite{xu2022reliable}.
In contrast, benefiting from large-scale pre-training, segment anything model (SAM)~\cite{kirillov2023segment} exhibits exceptional performance and generalization ability in natural image segmentation, which inspires us to employ SAM for single-frame shadow detection.
Nonetheless, SAM generally considers shadows as background.
Therefore, applying SAM to shadow detection poses a non-trivial challenge, as the main issue lies in bridging the gap between natural objects and complex shadows.
To unleash the potential of SAM for shadow detection, we conduct a detailed analysis and propose a fine-tuning strategy.
Specifically, SAM consists of three components: an image encoder, a prompt encoder, and a mask decoder. 
The image encoder trained on SA-1B~\cite{kirillov2023segment} can extract excellent representations of natural images and the prompt encoder is trained well and does not require further adjustments, so we freeze them. 
The mask decoder is lightweight (about 4M parameters), and we consider fine-tuning it.
Fine-tuning SAM requires the user to provide sparse or dense prompts to the prompt encoder.
In this paper, we simulate user interaction by utilizing bounding boxes obtained from the ground truth masks of the shadow locations (for more details on extraction of boxes, please refer to~\ref{how_to_finetune}).
Our fine-tuning strategy is simple yet effective, and the fine-tuned model~\cite{kirillov2023segment} is capable of detecting shadows more accurately with little user interaction that the user only needs to annotate the bounding boxes indicating the shadow position in the first frame of the video.
Subsequently, we utilize this detected shadow mask as a robust shadow prior to facilitate the detection of shadows in other frames of the video.

After obtaining the single-frame mask in the video, the next step is to propagate it across frames.
Attention mechanisms have been demonstrated effective in aggregating temporal and spatial information and have been widely utilized in the fields of video object segmentation and tracking~\cite{gao2022aiatrack,yang2021associating,mao2021joint,cheng2022xmem}. 
However, unlike the objects of interest with regular shapes in tracking tasks, shadows undergo significant deformations over time, making it difficult to track them consistently~\cite{liu2023scotch}.
To alleviate this issue, we combine self-attention and long short-term attention mechanisms to improve the alignment and propagation of fine-tuned SAM results across frames.
Specifically, the self-attention mechanism captures internal correlations of image representations, enabling better shadow detection results in the current frame.
The long-term attention mechanism establishes connections between the current and previous frames, consolidating spatial shadow information with past features.
The short-term attention mechanism aggregates adjacent frames in a spatial-temporal neighborhood with the current frame, resulting in smooth and continuous predictions of nearby frames.
The synergistic integration of long-term and short-term attention mechanisms offers an effective solution to detect shadows, yielding accurate and temporally coherent video shadow detection results.
In this paper, we consider the first frame of the video as the long-term frame and employ the last frame of the current frame as the short-term frame.

Formally, as illustrated in Figure~\ref{fig:intro}, we present ShadowSAM, a new framework for video shadow detection.
It comprises of two components: the fine-tuned SAM and the long short-term network that incorporates long short-term attention blocks.
Initially, we employ the fine-tuned SAM to generate the mask of the first frame as the shadow prior in the video.
Subsequently, we integrate the shadow prior with image features for alignment and propagation within the long short-term network and continuously update them with new shadow mask to facilitate subsequent shadow predictions.
% obtain key-value pairs which containing image features and shadow mask information and continuously update them with new shadow mask during the process of video prediction.
% These key-value pairs are utilized for alignment and propagation within the long short-term network and the newly generated pairs can facilitate subsequent shadow predictions.
We conduct comprehensive experiments on the ViSha dataset~\cite{chen2021triple} and validate the effectiveness of our proposed method.

In summary, our method has three contributions:

\begin{itemize}
	\item We introduce ShadowSAM, a new framework that incorporates a single-frame mask into video shadow detection, thereby enhancing detection performance.
	\item We employ a simple yet effective approach to fine-tune the lightweight mask decoder in SAM, enabling its successful application to shadow detection.
	\item We incorporate the long-term and the short-term attention into ShadowSAM for video shadow detection, which achieves promising performance while ensuring a fast inference speed compared to other techniques.
\end{itemize}

%%%%%%%%%%%%%%%%%%%%%%%%%%%% related work %%%%%%%%%%%%%%%%%%%%%%%%%%%%
\section{Related Work}
\subsection{Video Shadow Detection}
Single-image shadow detection is a common computer vision task that has been explored for many years~\cite{hu2018direction,zheng2019distraction,chen2020multi,zhu2021mitigating,hu2021revisiting,inoue2020learning,jie2023rmlanet}.
Inoue \emph{et al.}~\cite{inoue2020learning} propose a method for shadow image synthesis.
They extend the physically-grounded shadow illumination model and synthesize shadow images under various combinations of input elements, including shadow-free images, matte images, and shadow parameters. 
Ultimately, they present a large-scale synthetic dataset termed SynShadow. 
Experimental results conducted on this dataset demonstrate the effectiveness of their method.
Jie \emph{et al.}~\cite{jie2023rmlanet} introduce a random multi-level attention network, RMLANet.
This network effectively incorporates multi-level features and guiding features using self-attention mechanism.
Additionally, they propose a sparse attention mechanism to reduce the number of attention pairs, thereby decreasing computational complexity.
This method achieves notable results in single-image shadow detection.
In comparison, we mainly address the more challenging task of video shadow detection. 
While videos can provide temporal information, they also pose significant challenges.
For example, we should consider factors like shadow movement, inter-frame contextual information, and real-time constraints.

Video shadow detection aims to detect shadows in each frame of a video.
Traditional methods utilize hand-crafted features to predict shadow masks in videos, \emph{e.g.}, illumination~\cite{nadimi2004physical,stander1999detection}, intensity~\cite{elgammal2002background,jacques2005background}, color~\cite{cucchiara2003detecting,kumar2002comparative,salvador2004cast}, and others~\cite{xu2005insignificant,wang2009real,liu2011cast}.
Specifically, Nadimi \emph{et al.}~\cite{nadimi2004physical} propose a method based on a new spatio-temporal albedo test and dichromatic reflection model, which can address the problem of separating moving cast shadows from the moving objects without relying on any geometric assumptions.
Jacques \emph{et al.}~\cite{jacques2005background} improve the intensity-based background subtraction technique and propose a new method for detecting shadows in grayscale video sequences.
The proposed method works well in both indoor and outdoor scenarios.
Salvador \emph{et al.}~\cite{salvador2004cast} assume that the presence of shadows darkens the surface which they are cast upon and utilize the color invariance and geometric properties of shadows to detect shadows.
The proposed method is robust and efficient for a large class of scenes.
Xu \emph{et al.}~\cite{xu2005insignificant} first utilize an initial mask of cast shadows and generate a canny edge map.
Subsequently, the shadow region is detected through multiframe integration, edge matching, and region growing.
Russell \emph{et al.}~\cite{russell2017feature} propose a feature-based image patch approximation and multi-independent sparse representation technique to distinguish shadow points from the foreground object in many problematic situations.
This method constructs over-complete dictionaries for objects and shadows in new image patches by introducing two types of illumination-invariant features and intensity-reduction histogram, which are ultimately employed for patch matching.
However,  due to the intricacies of real-world scenes, hand-crafted features are insufficient in handling shadows, resulting in poor performance in video shadow detection.
In contrast, we devise a deep learning-based method to address the issue of video shadow detection, which can automatically extract features from images.
Additionally, deep learning-based methods can leverage pre-trained models to learn generic visual representations, further enhancing their capabilities in video shadow detection.

Recently, deep learning-based methods have made significant progress in video shadow detection. 
Chen \emph{et al.}~\cite{chen2021triple} first propose the large-scale video shadow detection dataset ViSha and a new framework TVSD-Net, which can simultaneously learn discriminative feature representations at intra-video and inter-video. 
Ding \emph{et al.}~\cite{ding2022learning} propose SC-Cor to learn the correspondence between shadows in consecutive frames, making the predicted shadow masks more temporally coherent. 
Lu \emph{et al.}~\cite{lu2022video} propose a spatio-temporal interpolation consistency pipeline to alleviate the issue of high generalization error and temporal inconsistent results.
They utilize both unlabeled video frames and labeled images to train the network and perform spatial and temporal interpolation to enhance the performance of video shadow detection.
More recently, Liu \emph{et al.}~\cite{liu2023scotch} introduce SCOTCH and SODA, a new framework to address the issue of large shadow deformations.
Specifically, SODA is designed to handle significant shadow deformations in videos, while SCOTCH is used to guide the network in learning a unified representation from positive shadow pairs across different videos. 
The proposed approach can effectively address the shadow deformation issues and learn better shadow representations in different videos.

In contrast to the previous methods, we utilize the visual foundation model SAM~\cite{kirillov2023segment} to provide prior knowledge, \emph{i.e.}, shadow mask from the first frame.
Based on this, we design an attention network to detect shadows in videos.

\begin{figure*}[t]
	\centering
	\includegraphics[width=0.98\linewidth]{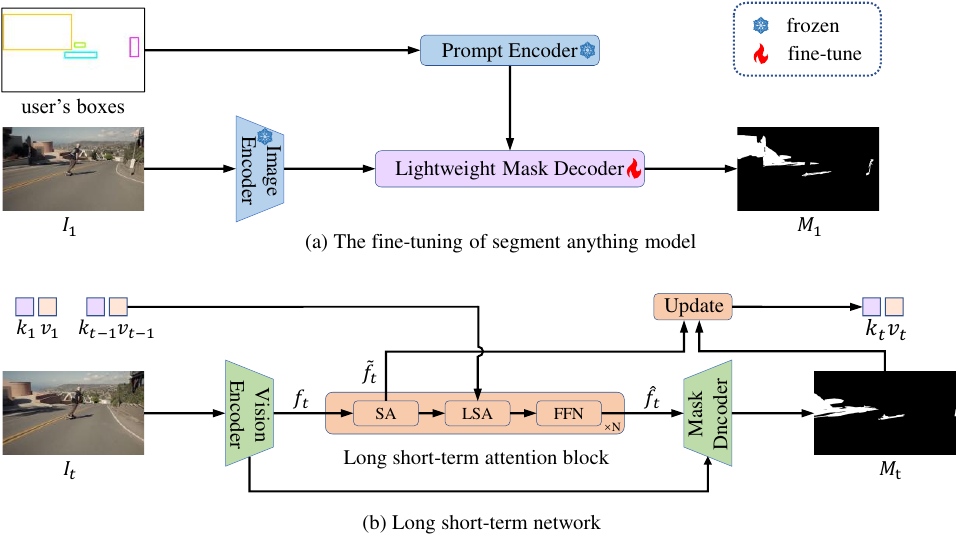}
	\caption{The overall architecture of the proposed ShadowSAM. It consists of two parts: the fine-tuned SAM and the long short-term network. The SAM utilizes the fine-tuned mask decoder to predict the shadow mask for the first frame of the video. The long short-term network propagates this initial mask across the entire video and incorporates the key-value pairs of previous frame through the long short-term attention mechanism to further improves the detection performance. The SA, LSA, and FFN in the long short-term attention block represent the self-attention layer, the long short-term attention layer, and the feed-forward network, respectively.}
	\label{fig:arch}
\end{figure*}

\subsection{Attention Mechanism}
In recent years, the attention mechanism has played an important role in deep learning techniques.
In the field of natural language processing (NLP), the attention mechanism in transformer~\cite{vaswani2017attention} has demonstrated remarkable performance in many downstream tasks.
The main component of transformer is the multi-head self-attention, which takes a sequence of embeddings as input and leverages linear projection and matrix multiplication operations to model their correlations. 
By introducing attention mechanism, the network effectively captures the interdependencies among sequences, thereby intensifying its focus on crucial sequence embeddings.
Inspired by the success of the attention mechanism~\cite{vaswani2017attention} in NLP, Vision Transformer (ViT)~\cite{dosovitskiyimage} first introduces transformer to the visual domain and proves its effectiveness.
Subsequently, the attention mechanism has witnessed increasing adoption in various computer vision tasks, \emph{e.g.}, DETR~\cite{carion2020end} for object detection and VisTR~\cite{wang2021end} for video instance segmentation.
The attention mechanism in the transformer has gained great popularity and emerged as a competitive framework that comparable to convolutional networks across various tasks~\cite{carion2020end,zhu2021deformable,dosovitskiyimage,liu2021swin}.

In this paper, considering the complex deformations of shadows, we combine self-attention mechanism with long-term and short-term attention~\cite{yang2021associating,cheng2022xmem,gao2022aiatrack} to detect shadows. 
The long-term attention mechanism captures the spatial position and quantity information of shadows in the video, while the short-term attention mechanism smooths the continuity of shadow mask between adjacent frames. 
These two mechanisms complement each other and assist the network to effectively perform video shadow detection.

% In the field of video, long and short-term attention are considered as dynamic information aggregation units that simultaneously learn spatial and temporal information~\cite{mao2021joint,yang2021associating,cheng2022xmem,yan2021learning,gao2022aiatrack}.
% Such attention mechanisms are widely used in video object segmentation~\cite{mao2021joint,yang2021associating,cheng2022xmem} and video object tracking~\cite{yan2021learning,gao2022aiatrack}.
% In this paper, we combine the long-term and short-term attention mechanism with self-attention layers to ensure that the detected shadows have both spatial consistency and temporal continuity.

\subsection{Foundation Models}
A foundation model refers to a model that is pre-trained (usually using self-supervised learning) on a wide range of datasets and can be quickly applied to downstream tasks through fine-tuning or in-context learning mechanisms.
One difference between the foundation model and the small model is that the intelligence of the foundation model gradually emerges as the amount of training data and model parameters increases~\cite{bommasani2021opportunities}.
Currently, foundation models have become mature in the field of NLP, \emph{e.g.}, BERT~\cite{devlin2019bert}, GPT-2~\cite{radford2019language}, ToBERTa~\cite{liu2019roberta}, and T5~\cite{raffel2020exploring}, which adopt the paradigm of self-supervised training and exhibit superior performance in contextual understanding.
In addition, models such as CLIP~\cite{radford2021learning} and BEiT-3~\cite{wang2022image} successfully bridge the relationship between multimodalities, aligning them semantically and making the models more standardized.
Segment anything~\cite{kirillov2023segment} first proposes a foundation model in the computer vision field for natural image segmentation, which is pre-trained on 11M images with 1B masks, and users can provide various types of prompts, \emph{e.g.}, points, boxes, masks, and texts, to enable the model to output the segmentation masks required by the user.
However, the segment anything model treats shadows as the background during the segmentation process and does not segment them.
To address this issue, we conduct an effective fine-tuning strategy to make SAM output precise shadow masks.

%%%%%%%%%%%%%%%%%%%%%%%%%%%% approach %%%%%%%%%%%%%%%%%%%%%%%%%%%%
\section{Approach}
\label{approach}
\subsection{An Overview of ShadowSAM}
Figure~\ref{fig:arch} illustrates the overall architecture of our framework.
Given a series of video frames $\bm{I}_{1}, \bm{I}_{2},...,\bm{I}_{T}$, we aim to utilize the fine-tuned SAM to predict the shadow mask $\bm{M}_{1}$ for the first frame $\bm{I}_{1}$ and then this mask is utilized for propagation to detect shadows in other frames of the video.

To obtain the shadow mask for the first frame, we employ bounding boxes as sparse prompts to fine-tune the SAM~\cite{kirillov2023segment}, which enables accurate shadow detection within the image.
Subsequently, we utilize a long short-term network to predict other frames sequentially.
Specifically, we fuse the self-attention feature $\tilde{\bm{f}}_{t}$ of frame $\bm{I}_{t}$ with the predicted mask $\bm{M}_{t}$ to obtain $\bm{k}_{t}$ and $\bm{v}_{t}$, which contain image feature and shadow information.
Note that for $\bm{k}_{1}$ and $\bm{v}_{1}$, we need to feed the first frame $\bm{I}_{1}$ into the long short-term network to extract features, which are then combined with the predicted shadow mask $\bm{M}_{1}$ from SAM to obtain them.
For frame $\bm{I}_{t+1}$, we utilize $\bm{k}_{1}$ and $\bm{v}_{1}$ as long-term features and $\bm{k}_{t}$ and $\bm{v}_{t}$ as short-term features for prediction.
The formulation is represented as follows:
\begin{equation}
	\begin{aligned}
		%		& \bm{v}_{t}, \bm{k}_{t} = \operatorname{Update}(\bm{f_{t}}, \bm{M}_{t}), \\
		& \bm{k}_{t}, \bm{v}_{t} = \operatorname{N_{update}}(\tilde{\bm{f}}_{t}, \bm{M}_{t}), \\
		& \bm{M}_{t+1} = \operatorname{N_{LSTN}}(\bm{I}_{t+1}, (\bm{k}_{1}, \bm{v}_{1}), (\bm{k}_{t}, \bm{v}_{t})), \\
	\end{aligned}
	\label{eq:overview}
\end{equation}
where $\operatorname{N_{update}}$ and $\operatorname{N_{LSTN}}$ are the update key-value pairs operation and the long short-term network, respectively. 
Here, we directly utilize the prediction of the fine-tuned SAM as the final result for the first frame $\bm{I}_{1}$.
For a new video, we repeat the above process.
In the following, we first revisit the SAM~\cite{kirillov2023segment} model in Section~\ref{revisit}. 
Then, we elaborate on how to obtain the bounding boxes from ground truth shadow masks to simulate user interaction and utilize these bounding boxes to fine-tune SAM for single-image shadow detection in Section~\ref{how_to_finetune}.
Finally, we introduce the details of the long short-term network in Section~\ref{longshort}.

%In the following, we first review the SAM~\cite{kirillov2023segment}, and then elaborate on how to fine-tune SAM and the details of long short-term network, respectively.

\subsection{Preliminaries}
\label{revisit}

\setlength{\tabcolsep}{8mm}
\begin{table}[t]
	\caption{Table of symbol definitions.}
	\centering
	\resizebox{0.9\linewidth}{!}{
		\begin{tabular}{ll}
			\Xhline{2.5\arrayrulewidth}
			Symbol & Meaning \\
			\hline
			\hline
			$\bm{I}$ & input image \\
			$\bm{P}$ & user prompt \\
			$\bm{f_{i}}$ & image embedding \\
			$\bm{f_{p}}$ & prompt embedding \\
			$E_{i}$ & image encoder \\
			$E_{p}$ & prompt encoder \\
			$E_{m}$ & mask decoder \\
			$\bm{f}_{o}$ & output token \\
			$\bm{M}$ & output mask \\
			% \hline
			\Xhline{2.5\arrayrulewidth}
		\end{tabular}
	}
	\label{tab:symbol}
\end{table}

Segment anything model (SAM)~\cite{kirillov2023segment} is pre-trained on 1B masks of 11M images, aiming to segment corresponding objects in natural images through user interaction.
The main inspiration for SAM comes from NLP, where foundation models are pre-trained through predicting tokens and then fine-tuned through prompt engineering for downstream tasks.
Before introducing SAM~\cite{kirillov2023segment}, we present the symbols that will be used in Table~\ref{tab:symbol} for better understanding and readability.
SAM~\cite{kirillov2023segment} consists of three main components: an image encoder $E_{i}$, a flexible prompt encoder $E_{p}$, and a lightweight mask decoder $E_{m}$.
Specifically, the image encoder uses an MAE~\cite{he2022masked} pre-trained Vision Transformer (ViT)~\cite{dosovitskiyimage}.
Given an image $\bm{I}$ of arbitrary size, it is first rescaled and padded to resolution $1024\times 1024$, and then fed to ViT to obtain an image embedding.
There are three variants of ViT, namely ViT-H, ViT-L, and ViT-B.
In this paper, we apply ViT-B to extract image features.
%In this paper, we apply ViT-B, which C, H, and W are 256, 64, and 64, respectively.
For the prompt encoder, there are two types of prompt methods: sparse (points, boxes, text) and dense (masks). 
Sparse prompts are encoded as positional embeddings and added to learnable positional embeddings, while dense prompts are added to image embeddings through operations such as convolution. 
In this paper, to fine-tune SAM more efficiently, we extract bounding boxes from the ground truth shadow masks as sparse prompts.
To simulate user interaction as closely as possible, we analyze the rules for extracting the bounding boxes and provide a detailed description in Section~\ref{how_to_finetune}.
For the mask decoder, it is utilized to aggregate features from the image encoder and the prompt encoder and produces the final output.

More specifically, given an image $\bm{I}$ and the user prompt $\bm{P}\in \mathbb{R}^{N\times 4}$, where $N$ is the number of bounding boxes, the process of image encoder and prompt encoder can be represented as follows,
\begin{equation}
	\bm{f}_{i} = E_{i}(\bm{I}), \bm{f}_{p} = E_{p}(\bm{P}),
	\label{eq:revisit_1}
\end{equation}
where $\bm{f_{i}}\in \mathbb{R}^{H\times W\times C}$ and $\bm{f_{p}}\in \mathbb{R}^{N\times 2\times C}$ are image embedding and prompt embedding, respectively, and $H$, $W$ and $C$ are set as 64, 64, and 256, respectively.
The mask decoder is lightweight and modified from a transformer block~\cite{vaswani2017attention}, followed by a dynamic mask prediction head.
It updates both the image embedding and the prompt embedding using self-attention and cross-attention mechanisms. 
In addition, this module introduces output tokens as dynamic mask prediction units, which are then fed into an MLP to calculate the probability of mask foreground at each position in the image.
This process can be expressed as follows:
\begin{equation}
	\bm{M} = E_{m}(\bm{f}_{i}, \bm{f}_{p}, \bm{f}_{o}),
	\label{eq:revisit_2}
\end{equation}
where $\bm{f}_{o}\in \mathbb{R}^{5\times C}$ and $\bm{M}$ are output token and final mask, respectively.
In this paper, we freeze the image encoder and prompt encoder, and only fine-tune the lightweight mask decoder to enable effective shadow detection.

\subsection{Fine-tuning of Segment Anything Model}
\label{how_to_finetune}
In this section, we present fine-tuning of segment anything model~\cite{kirillov2023segment} for shadow detection, which is divided into two parts. 
Firstly, we introduce the overall fine-tuning strategy of SAM~\cite{kirillov2023segment}.
Then we analyze how to generate bounding boxes from the ground truth shadow masks as sparse prompts for SAM to better simulate user interaction.

\begin{figure}[t]
	\centering
	\includegraphics[width=0.98\linewidth]{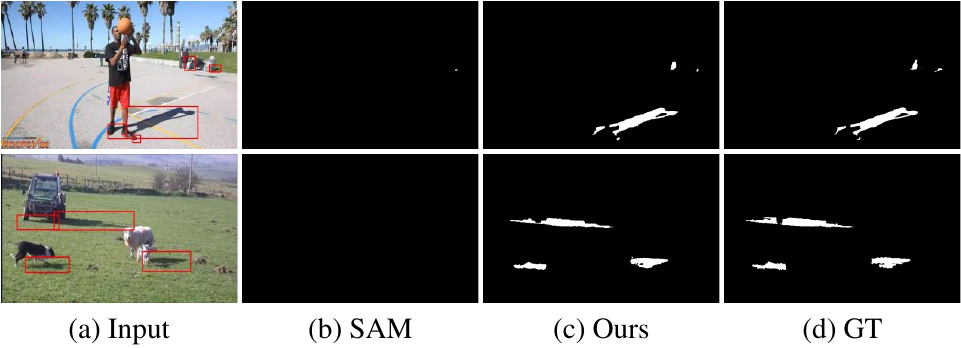}
	\caption{Visual comparison of our fine-tuned SAM with the original SAM, it can be concluded that the original SAM treats the shadow as background, resulting in no output.}
	\label{fig:finesam}
\end{figure}

\noindent
\textbf{Fine-tuning strategy.}
As mentioned above, our fine-tuning strategy focuses on fine-tuning the mask decoder in SAM~\cite{kirillov2023segment} on the ViSha dataset~\cite{chen2021triple}.
Our rationale is as follows: for the image encoder, it is pre-trained on millions of natural images using MAE~\cite{he2022masked}, and the trained encoder is capable of extracting good natural image representations, which are perfectly applicable for video shadow detection.
In addition, the image encoder accounts for the majority of the parameters and computational costs of the entire SAM network.
Fine-tuning it requires a significant amount of training data and is computational expensive.
%For the prompt encoder, we apply the sparse prompt, boxes, which are considered to have no gap to the current task, so we also freeze it.
For the prompt encoder, we also freeze it and apply bounding boxes as prompts.
Therefore, we only fine-tune the lightweight mask decoder (about 4M parameters).
Fine-tuning SAM requires incorporating prompt information.
We use the bounding boxes extract from the ground truth shadow masks as sparse prompts and feed them to the prompt encoder.
The process of obtaining the bounding boxes is described next.

\noindent
\textbf{Bounding boxes generation.}
Designing the bounding boxes for the shadow region is a critical problem that requires careful consideration.
There are two main challenges.
Firstly, shadows have complex shapes and appear at various locations in natural images due to the shapes of occluding objects, illumination angles, and projection positions.
Secondly, during the training, the designing of bounding boxes should simulate the interaction between users and the model as closely as possible, to better reflect practical application.
For the first challenge, we locate all eight-connected regions in the ground truth shadow mask and discard regions that contain fewer than 50 pixels.
Then, we extract the minimum bounding box of each effective region as the sparse prompt for the current image. 
If the number of boxes in the current image exceeds a threshold (set to 8 in this paper), we use the size of the whole image as the bounding box.
For the second, since users may not provide perfectly accurate boxes that fit the shadow region, we apply a random perturbation of 0$\sim$20 pixels to each of four boundaries of the boxes.
We utilize the obtained boxes as prompts to fine-tune SAM.
As shown in Figure~\ref{fig:finesam}, with boxes as prompts, the fine-tuned SAM can detect shadows effectively. 
In contrast, the original SAM treats shadows as background, leading to invalid shadow masks.

\begin{figure}[t]
	\centering
	\includegraphics[width=0.98\linewidth]{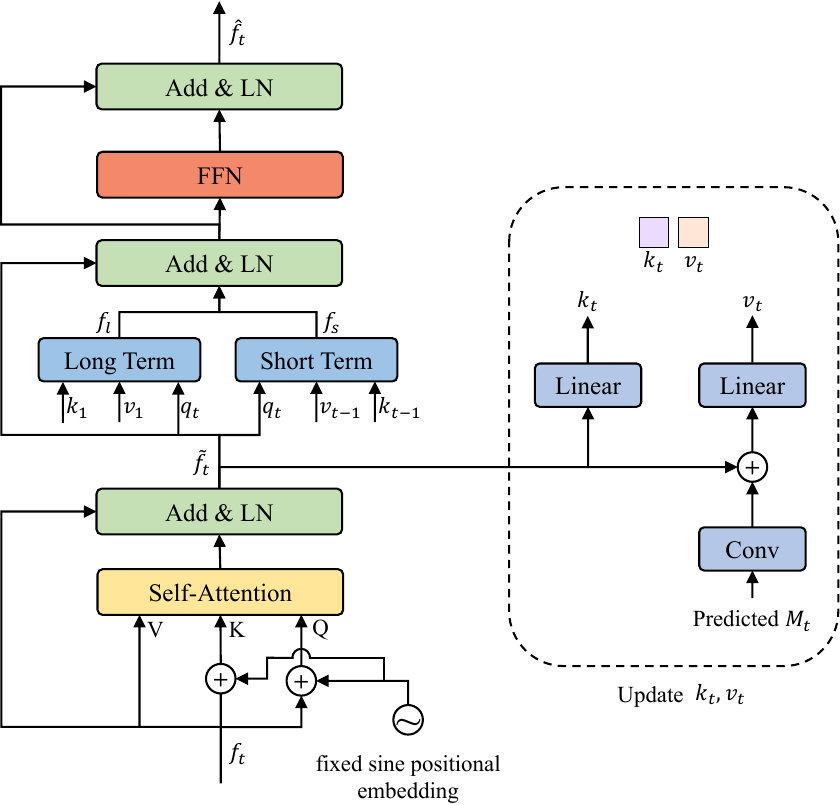}
	\caption{Long short-term attention block and the update module. We apply long short-term attention block to capture the spatial and temporal information simultaneously. After the process, we use the self-attention feature $\tilde{\bm{f}}_{t}$ to update key-value pairs $\bm{k}_{t}$ and $\bm{v}_{t}$ of current frame.}
	\label{fig:attn}
\end{figure}

\subsection{Long Short-Term Network}
\label{longshort}
The long short-term network consists of an image encoder, several long short-term attention blocks, and a mask decoder.
We apply a lightweight backbone MobileNetV2~\cite{sandler2018mobilenetv2} as the image encoder to extract features and FPN~\cite{lin2017feature} as mask decoder.
The MobileNetV2~\cite{sandler2018mobilenetv2} is employed to extract features from the input images, which are subsequently fed into the long short-term attention block (LSAB).
LSAB is similar to the long-term and short-term attention in AOT~\cite{yang2021associating} that we combine the self-attention mechanism with long short-term attention to better learn the correlation between previous frames and the current frame.
Firstly, the long short-term attention block learns the features of shadows within the current frame through a self-attention layer.
Then, long-term and short-term attention are performed to establish spatial object correlations and temporal object continuity. 
Finally, the aggregated features are output through two-layer MLP with GroupNorm~\cite{hendrycks2016gaussian} inserted in the middle to obtain the optimized features, which are fed into the FPN~\cite{lin2017feature} decoder to obtain the shadow mask.
In addition, after processing current frame in the video, we update the key-value pairs using the current frame features and the predicted mask so as to serve the prediction of subsequent frames.

\noindent
\textbf{Key-value update.}
As shown in Figure~\ref{fig:attn}, the feature map $\bm{f}_{t}\in \mathbb{R}^{H\times W\times C}$ of frame $\bm{I}_{t}$ generated by the image encoder is transformed into the feature $\tilde{\bm{f}}_{t}\in \mathbb{R}^{H\times W\times C}$ through a self-attention layer, where $H$, $W$, and $C$ are set as 30, 30, and 256, respectively.
This feature is reshaped and fed into a linear transformation function $\phi$ to derive $\bm{k}_{t}\in \mathbb{R}^{HW\times C}$. 
Subsequently, the predicted shadow mask $\bm{M}_{t}$ of frame $\bm{I}_{t}$, which has a spatial correlation with the shadow feature, is passed to a convolutional operation to match the resolution of $\tilde{\bm{f}}_{t}\in \mathbb{R}^{H\times W\times C}$ and added to it.
Then, another linear transformation function $\psi$ is applied to obtain $\bm{v}_{t}\in \mathbb{R}^{HW\times C}$.
The whole progress can be expressed as follows:
\begin{equation}
	\begin{aligned}
		& \bm{k}_{t} = \phi(\tilde{\bm{f}_{t}}), \\
		& \bm{v}_{t} = \psi(\tilde{\bm{f}_{t}}+\operatorname{Conv}(\bm{M}_{t})). \\
	\end{aligned}
	\label{eq:memory}
\end{equation}
The two features $\bm{k}_{t}$ and $\bm{v}_{t}$ are subsequently applied to serve the prediction of subsequent frames.

\setlength{\tabcolsep}{7pt}
\begin{table*}[t]
	%	 \small
	%	 \footnotesize
	\caption{Quantitative comparison with previous works on ViSha dataset~\cite{chen2021triple}. The best and the second results are highlighted in \textbf{bold} and \underline{underlined}, respectively. ``$\uparrow$'' indicates the higher the better and ``$\downarrow$'' means the opposite. SBER and NBER indicate the BER value in shadow region and non-shadow region, respectively. SP, SOD, VOS, ISD, and VSD represent the scene parsing, salient object detection, video object segmentation, single-image shadow detection, and video shadow detection methods, respectively.}
	\centering
	\begin{tabular}{l|l|c|cccccc}  
		\Xhline{2.5\arrayrulewidth}
		%		\hline
		\textbf{Method} & \textbf{Task} & \textbf{Venue} & \textbf{MAE}$\downarrow$  & \textbf{$\bm{F}_{\beta}$}$\uparrow$  & \textbf{IoU}$\uparrow$ & \textbf{BER}$\downarrow$ & \textbf{SBER}$\downarrow$ & \textbf{NBER}$\downarrow$  \\
		\hline
		\hline
		FPN~\cite{lin2017feature}    & SP  &  \emph{CVPR'17}  &  0.044  &  0.707  &  0.513   &  19.49  &  36.59  &  2.40   \\
		PSPNet~\cite{zhao2017pyramid} & SP &  \emph{CVPR'17}  &  0.052  &  0.642  &  0.476  &  19.75   &  36.43  &  3.07   \\
		\hline
		DSS~\cite{hou2017deeply}  & SOD &  \emph{CVPR'17}  & 0.045   & 0697  &  0.503  &  19.78  &  36.96  &  2.59  \\
		MGA~\cite{li2019motion}  & SOD &  \emph{ICCV'19}  & 0.068   & 0.591  &  0.398  &  25.77  &  48.17  &  3.37   \\
		\hline
		PDBM~\cite{song2018pyramid}  & VOS &  \emph{ECCV'18}  & 0.066  &  0.623  &  0.466  &  19.77   &  34.37  &  5.16  \\
		FEELVOS~\cite{voigtlaender2019feelvos} & VOS &  \emph{CVPR'19}  & 0.043  &  0.710   &  0.512   &  19.76  &  37.27  &  2.26   \\
		STM~\cite{oh2019video}         & VOS &  \emph{ICCV'19}  & 0.069   & 0.598  &  0.408  &  25.69  &  47.44  &  3.95   \\	
		COSNet~\cite{lu2019see}      & VOS &  \emph{CVPR'19}  & 0.040  & 0.706  &   0.515  &   20.51   &  39.22  &  1.792  \\				               
		\hline
		BDRAR~\cite{zhu2018bidirectional} & ISD  &  \emph{ECCV'18}  & 0.050  & 0.695  & 0.484 &  21.30  &  40.28  &  2.32    \\
		DSD~\cite{zheng2019distraction} & ISD &  \emph{CVPR'19}  & 0.044  & 0.702  & 0.519  &  19.89  &  37.88  &  1.89  \\	
		MTMT~\cite{chen2020multi}       & ISD  &  \emph{CVPR'20}  & 0.043  & 0.729  & 0.517  &  20.29  &  38.71  &  1.86   \\	
		FSD~\cite{hu2021revisiting}   & ISD      &  \emph{TIP'21}  & 0.057   & 0.671  & 0.486 &  20.57  &  38.06  &  3.06  \\	
		\hline
		TVSD~\cite{chen2021triple} & VSD  &  \emph{CVPR'21}  & 0.033   &  0.757  & 0.567 &  17.70  &  33.97  &  1.45   \\
		%		& Hu  \emph{et al.}~\cite{hu2021temporal} & 2021 & 0.078 & 0.683 & 0.510 & 17.03 & 30.13 & 3.93  \\
		STICT~\cite{lu2022video}  & VSD   & \emph{CVPR'22} & 0.046 & 0.702 & 0.545 & 16.60 & 29.58 & 3.59  \\
		SC-Cor~\cite{ding2022learning}  & VSD  & \emph{ECCV'22} & 0.042 & 0.762 & 0.615 & 13.61 & 24.31 & 2.91 \\
		Liu \emph{et al.}~\cite{liu2023scotch} & VSD   & \emph{CVPR'23} & \underline{0.029} & \underline{0.793} & \underline{0.640} & \textbf{9.066} & \textbf{16.26} & \underline{1.44}  \\
		Ours                        & VSD  & -     & \textbf{0.020}     &  \textbf{0.816}     & \textbf{0.693}    &  \underline{11.35}    &  \underline{21.75}     &  \textbf{0.93}   \\
		\Xhline{2.5\arrayrulewidth}
	\end{tabular}
	\label{tab:qua}
	%	\vspace{-0.25in}
\end{table*}

\noindent
\textbf{Long-term attention.}
The long-term attention is used to model the spatial object information between the previous frame and the current frame. 
In this paper, we use the first frame as the long-term frame.
As time goes by, the distance between the current frame and the first frame will gradually increase, leading to error drifts~\cite{oh2019video}.
The introduction of long-term attention allows for the incorporation of spatial object numbers information, assisting the network in mitigating the issue of lost shadow information during shadow propagation.
As shown in Figure~\ref{fig:attn}, given the feature $\bm{f}_{t}\in \mathbb{R}^{H\times W\times C}$ of the frame $\bm{I}_{t}$, we can obtain feature $\tilde{\bm{f}}_{t}\in \mathbb{R}^{H\times W\times C}$ through self-attention and LayerNorm~\cite{ba2016layer}. 
Then we obtain $\bm{q}_{t}\in \mathbb{R}^{HW\times C}$ by feeding $\tilde{\bm{f}}_{t}\in \mathbb{R}^{H\times W\times C}$ to the same linear transformation $\phi$ as $\bm{k}_{1}\in \mathbb{R}^{HW\times C}$.
The formulation for long-term attention can be expressed as follows:
%\begin{equation}
%	\bm{f}_{l} = \operatorname{Attn_{long}}(\bm{q}_{t}, \bm{k}_{1}, \bm{v}_{1}),
%	\label{eq:long}
%\end{equation}
\begin{equation}
	\bm{f}_{l} = \operatorname{Attn_{long}}(\bm{q}_{t}, \bm{k}_{1}, \bm{v}_{1}),
	\label{eq:long}
\end{equation}
where $\bm{f}_{l}\in \mathbb{R}^{H\times W\times C}$ is the output of the long-term feature.

\noindent
\textbf{Short-term attention.}
The short-term attention is used to capture temporal information between adjacent frames and produces temporally consistent shadow masks.
In this paper, we use  the last frame of current frame as the short-term frame. 
When processing consecutive frames, we use the feature extracted from the last frame as the short-term memory to update feature of the current frame.
Since the two frames are close, the shadows will not undergo significant movements.
Therefore, in short-term attention, the feature vector of current frame at location $p$ interacts only with the features of the last frame within a spatial neighborhood of $w\times w$ around it.
%there will not be significant changes between the features, and thus the feature of the current frame interacts only with a nearby spatial-temporal neighborhood of the last frame. 
%Specifically, the feature at location $p$ interacts only with the features within a spatial neighborhood of $w\times w$ around it.
The formula for short-term attention is expressed as follows,
\begin{equation}
	\bm{f}_{s} = \operatorname{Attn_{short}}(\bm{q}_{t}, \bm{k}_{t-1}, \bm{v}_{t-1}),
	\label{eq:short}
\end{equation}
where $\bm{f}_{s}\in \mathbb{R}^{H\times W\times C}$ is the output of short-term feature and $w$ is set as 15.
Then, we add the long-term feature $\bm{f}_{l}\in \mathbb{R}^{H\times W\times C}$ and short-term feature $\bm{f}_{s}\in \mathbb{R}^{H\times W\times C}$ to the self-attention feature $\tilde{\bm{f}}_{t}\in \mathbb{R}^{H\times W\times C}$ to obtain better representation.
Next, the features derived from the long short-term attention module are input to the mask decoder for final prediction.

\subsection{Implementation Details}
\noindent
\textbf{Loss function.}
In this section, we summarize our loss function employed in ShadowSAM.
For fine-tuning SAM, we utilize cross-entropy loss~\cite{nowozin2014optimal} to optimize the lightweight mask decoder of SAM~\cite{kirillov2023segment} in the training process.
While for long short-term network, the loss function is an unweighted sum between cross-entropy loss and soft Jaccard loss~\cite{nowozin2014optimal,rahman2016optimizing}.

\noindent
\textbf{Training details.}
We implement ShadowSAM on the Linux platform using PyTorch with NVIDIA GTX 3090Ti GPUs.
We fine-tune the original SAM~\cite{kirillov2023segment} for 20 epochs using the ViSha training set~\cite{chen2021triple} with a learning rate of 1e-4.
While for the long short-term network, we apply a lightweight backbone MobileNetV2~\cite{sandler2018mobilenetv2} and FPN~\cite{lin2017feature} as the image encoder and mask decoder, respectively.
The backbone network is pre-trained on ImageNet~\cite{deng2009imagenet}, while others are trained from scratch.
We set the mini-batch size to 16 for training 20000 steps.
The input image is randomly cropped and resized to 465 for data augmentation.
The Adam~\cite{kingma2014adam} optimizer with a weight decay of 0.07 is used to optimize the whole network, and the learning rate is set to 2e-5 and 2e-4 for the pre-trained layers and scratch layers, respectively.

\noindent
\textbf{Inference.}
We apply the fine-tuned SAM~\cite{kirillov2023segment} to predict the shadow mask of the first and last frames in a video.
For the practical application, it is required for users to provide bounding boxes to SAM~\cite{kirillov2023segment}. 
In this paper, to expedite the performance evaluation of our method , we extract bounding boxes from the ground truth shadow masks and perturb them in experiments to simulate user-provided bounding boxes.
% The method used to extract the bounding boxes in these two frames is consistent with the training phase that they are derived from the ground truth shadow masks.
Then, we use the two frames and long short-term network to perform forward and backward predictions to obtain two results.
For frames whose IoU between two predictions is lower than 75\%, we apply the fine-tuned SAM to re-predict them.

%%%%%%%%%%%%%%%%%%%%%%%%%%%% experiment %%%%%%%%%%%%%%%%%%%%%%%%%%%%
\section{Experiment}
\label{exp}
\subsection{Training Dataset and Metrics}
\noindent
\textbf{Datasets.}
We validate and test our method on the ViSha dataset~\cite{chen2021triple}. 
ViSha is the first learning-oriented dataset built for video shadow detection. 
It contains 120 videos with a total of 11,685 frames over 390 seconds, which are divided into 50 training videos and 70 testing videos.
Each frame is annotated with a high-quality pixel-level shadow mask.
Additionally, the dataset exhibits diversity in shadow object types, quantities, and environmental conditions.

\noindent
\textbf{Metrics.}
Following previous methods~\cite{chen2021triple,ding2022learning,hu2018direction}, we use four commonly used validation metrics to evaluate the effectiveness of the method, including Mean Absolute Error (MAE)~\cite{chen2021triple}, F-measure ($F_{\beta}$)~\cite{chen2021triple,hu2021revisiting}, Intersection over Union (IoU)~\cite{chen2021triple}, and Balance Error Rate (BER)~\cite{chen2021triple,zhu2021mitigating}.
For MAE and BER, the lower the better, while for the metrics of F-measure and IoU, the higher are preferred.

\subsection{Comparison with the State-of-the-art Methods}
Video shadow detection has gained popularity recently, and there are a few related methods.
Therefore, following the previous methods~\cite{chen2021triple,lu2022video,ding2022learning,liu2023scotch}, we compare our method with five different kinds of vision tasks: scene parsing (SP), salient object detection (SOD), video object segmentation (VOS), single-image shadow detection (ISD), and video shadow detection (VSD), including FPN~\cite{lin2017feature}, PSPNet~\cite{zhao2017pyramid}, DSS~\cite{hou2017deeply}, MGA~\cite{li2019motion}, PDBM~\cite{song2018pyramid}, FEELVOS~\cite{voigtlaender2019feelvos}, STM~\cite{oh2019video}, COSNet~\cite{lu2019see}, BDRAR~\cite{zhu2018bidirectional}, DSD~\cite{zheng2019distraction}, MTMT~\cite{chen2020multi}, FSD~\cite{hu2021revisiting}, TVSD~\cite{chen2021triple}, STICT~\cite{lu2022video}, SC-Cor~\cite{ding2022learning} and Liu \emph{et al.}~\cite{liu2023scotch}.
We re-train their model on the ViSha dataset~\cite{chen2021triple} using publicly available code or obtain shadow detection results directly from their papers or open-source repositories, \emph{e.g.}, TVSD~\cite{chen2021triple}.
We conduct both quantitative and qualitative analyses of the results.
Further, we compare the computational efficiency of current video shadow detection methods, as described below.

\begin{figure*}[t]
	\centering
	\includegraphics[width=0.98\linewidth]{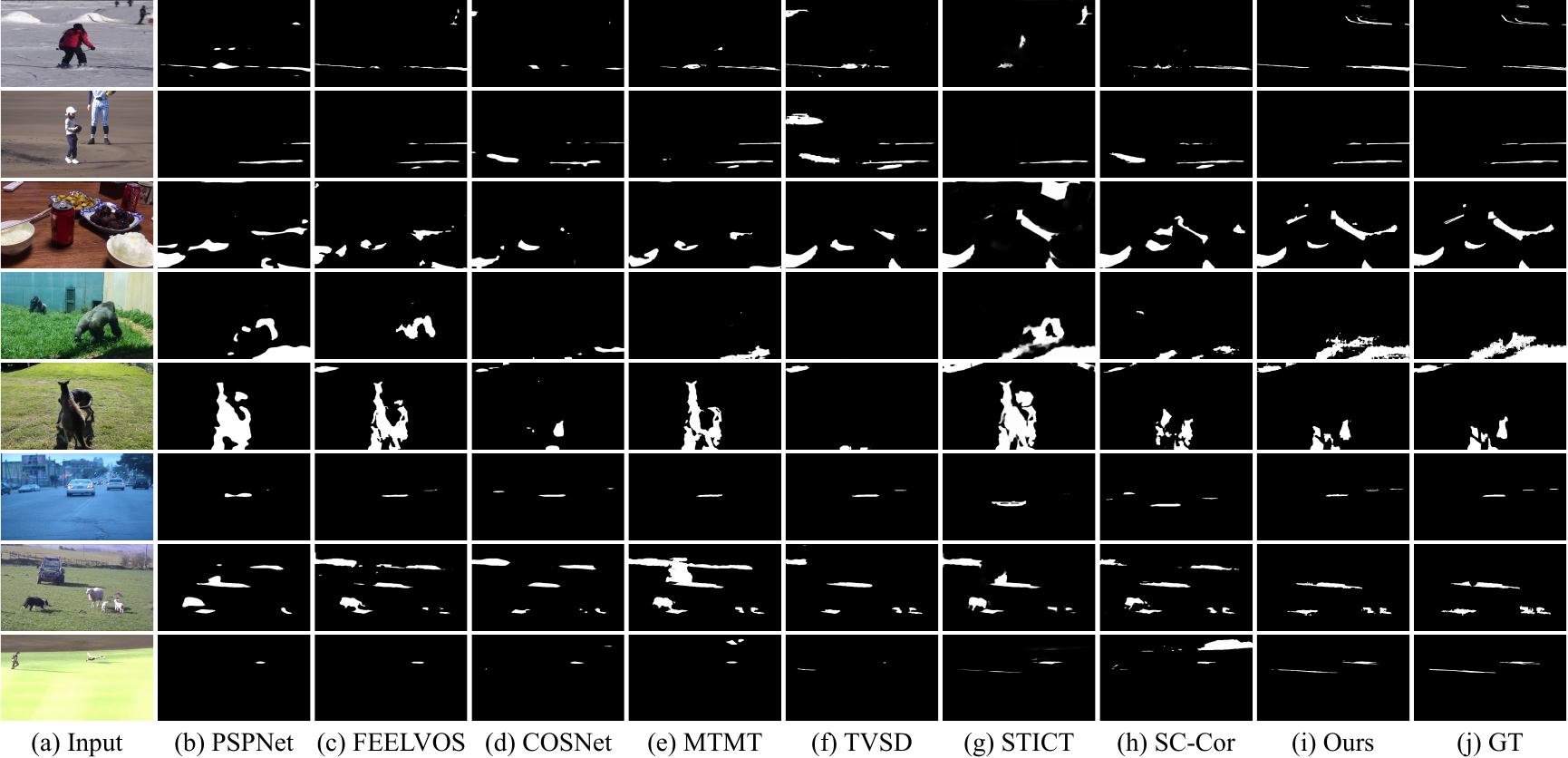}
	\caption{Visual comparison results of video shadow detection produced by different methods. (b) to (h) are results from SOTA methods: PSPNet~\cite{zhao2017pyramid}, FEELVOS~\cite{voigtlaender2019feelvos}, COSNet~\cite{lu2019see}, MTMT~\cite{chen2020multi}, TVSD~\cite{chen2021triple}, STICT~\cite{lu2022video}, and SC-Cor~\cite{ding2022learning}, respectively.}
	\label{fig:qua1}
\end{figure*}

\begin{figure}[t]
	\centering
	\includegraphics[width=1.\linewidth]{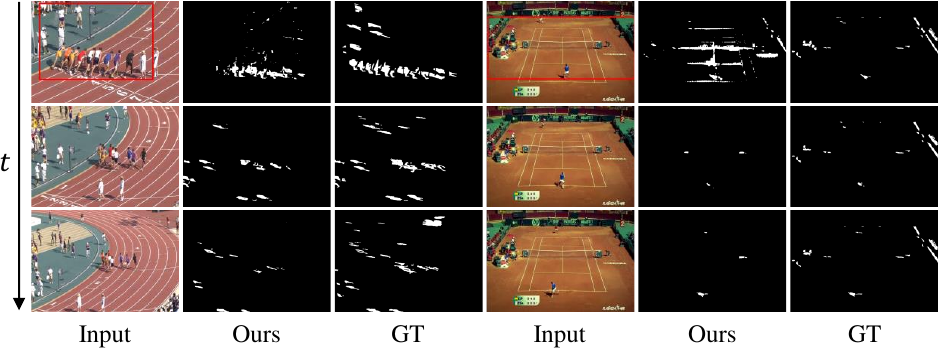}
	\caption{Failure cases. The first row represents the first frame of the video, while the second and third rows represent other frames of the video. When the video contains more than eight small objects, we employ the largest bounding box encompassing all shadows as user interaction. For such a coarse bounding box, the fine-tuned SAM struggles to produce an ideal shadow mask, which leads to the loss of some object shadows during the propagation of long short-term network.}
	\label{fig:neg}
\end{figure}

\noindent
\textbf{Quantitative comparison results.}
Table~\ref{tab:qua} summarizes the quantitative comparison results between our method and other state-of-the-art methods in terms of MAE, $F_{\beta}$, IoU, and BER.
For the methods of SP, SOD, and ISD, as they are single-image prediction tasks, they are unable to capture temporal information in videos during training.
Although they achieve acceptable performance in various metrics, they still have limitations.
For the VOS methods, they are designed to handle video tasks, which take into account temporal information compared to single-image methods.
However,  the complexity of shadows in terms of quantity and shape restricts their applicability.
In contrast, our method achieves better quantitative results.
Compared to the VSD methods, our approach achieves good results in terms of MAE, $F_{\beta}$, IoU, and BER metrics. 
Specifically, we improve the MAE by 31.0\% compared to TVSD~\cite{chen2021triple}. 
Moreover, we increase the $F_{\beta}$ from 0.762 to 0.816 and the IoU from 0.615 to 0.693 compared to SC-Cor~\cite{ding2022learning}.
Compared to Liu \emph{et al.}~\cite{liu2023scotch}, we achieve state-of-the-art results in terms of MAE, $F_{\beta}$, and IoU metrics.
Regarding the BER in the whole image and shadow region, the performance of our method is slightly inferior to them. 
However, for the BER in non-shadow region, our method achieves the best performance.
We argue that this is attributed to the fact that SAM~\cite{kirillov2023segment}, pretrained on 11M images accompanied by 1B masks, is initially designed as a foundation model for segmenting natural objects, while for shadows, SAM~\cite{kirillov2023segment}  tends to consider them as background and does not perform segmentation on them.
In this paper, we fine-tune the SAM~\cite{kirillov2023segment} on the ViSha dataset~\cite{chen2021triple}.
The fine-tuned SAM still remains its inherent bias towards segmenting natural objects, which leads to an instability to accurately detect shadows within the bounding boxes, particularly in cases involving densely distributed small objects (for more details, please refer to Figure~\ref{fig:neg}).
In contrast, Liu \emph{et al.}~\cite{liu2023scotch} introduce the shadow deformation attention trajectory (SODA), which is specifically designed to address the issue of large shadow deformations.
Moreover, this module employs a trajectory attention method~\cite{patrick2021keeping} to track the shadow, which aggregates spatial-temporal information of the shadow and enables the detection of shadows with intricate shapes.
Therefore, our method demonstrates improved performance in non-shadow region, while exhibiting reduced performance in terms of BER in shadow region, resulting in an inferior BER performance in the whole image compared to Liu \emph{et al.}~\cite{liu2023scotch}. 
We point out that this issue could be alleviated by fine-tuning SAM with more shadow data, thereby further enhancing its performance.

% We also achieve competitive results of BER with them, and our method effectively reduces false detections in non-shadow regions.

\begin{figure*}[t]
	\centering
	\includegraphics[width=0.98\linewidth]{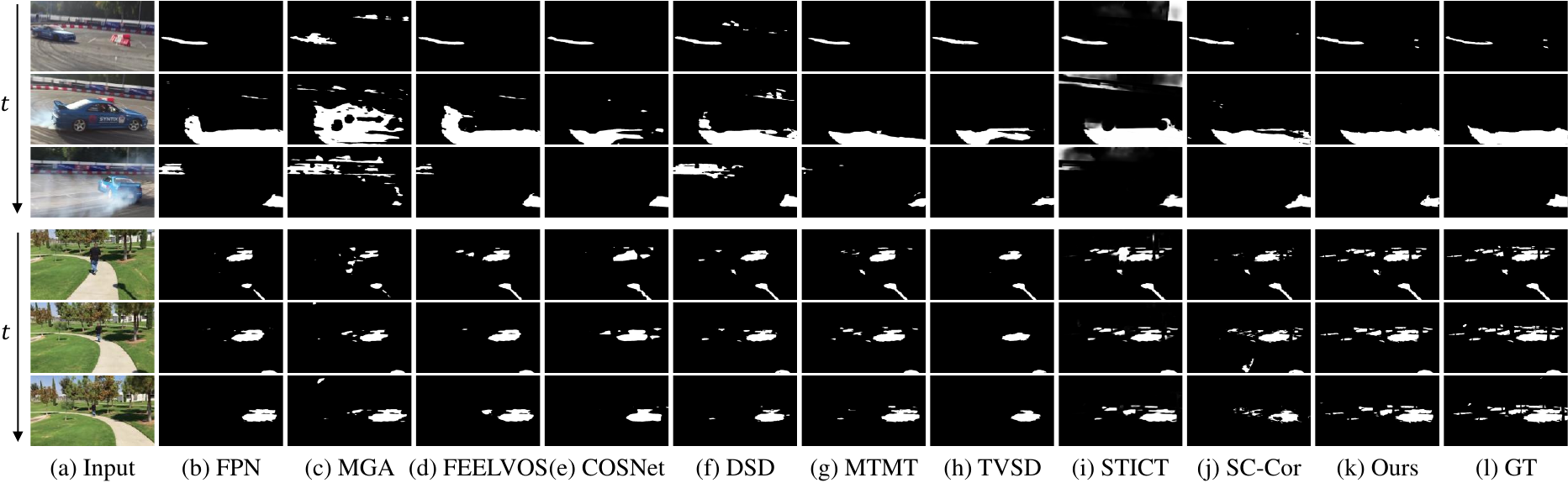}
	\caption{Visual comparison results of video shadow detection produced by different state-of-the-art methods. (b) to (j) are temporal prediction results from SOTA methods: FPN~\cite{lin2017feature}, MGA~\cite{li2019motion}, FEELVOS~\cite{voigtlaender2019feelvos}, COSNet~\cite{lu2019see}, DSD~\cite{zheng2019distraction}, MTMT~\cite{chen2020multi}, TVSD~\cite{chen2021triple}, STICT~\cite{lu2022video}, and SC-Cor~\cite{ding2022learning}, respectively.}
	\label{fig:qua2}
\end{figure*}

\begin{figure*}[t]
	\centering
	\includegraphics[width=0.98\linewidth]{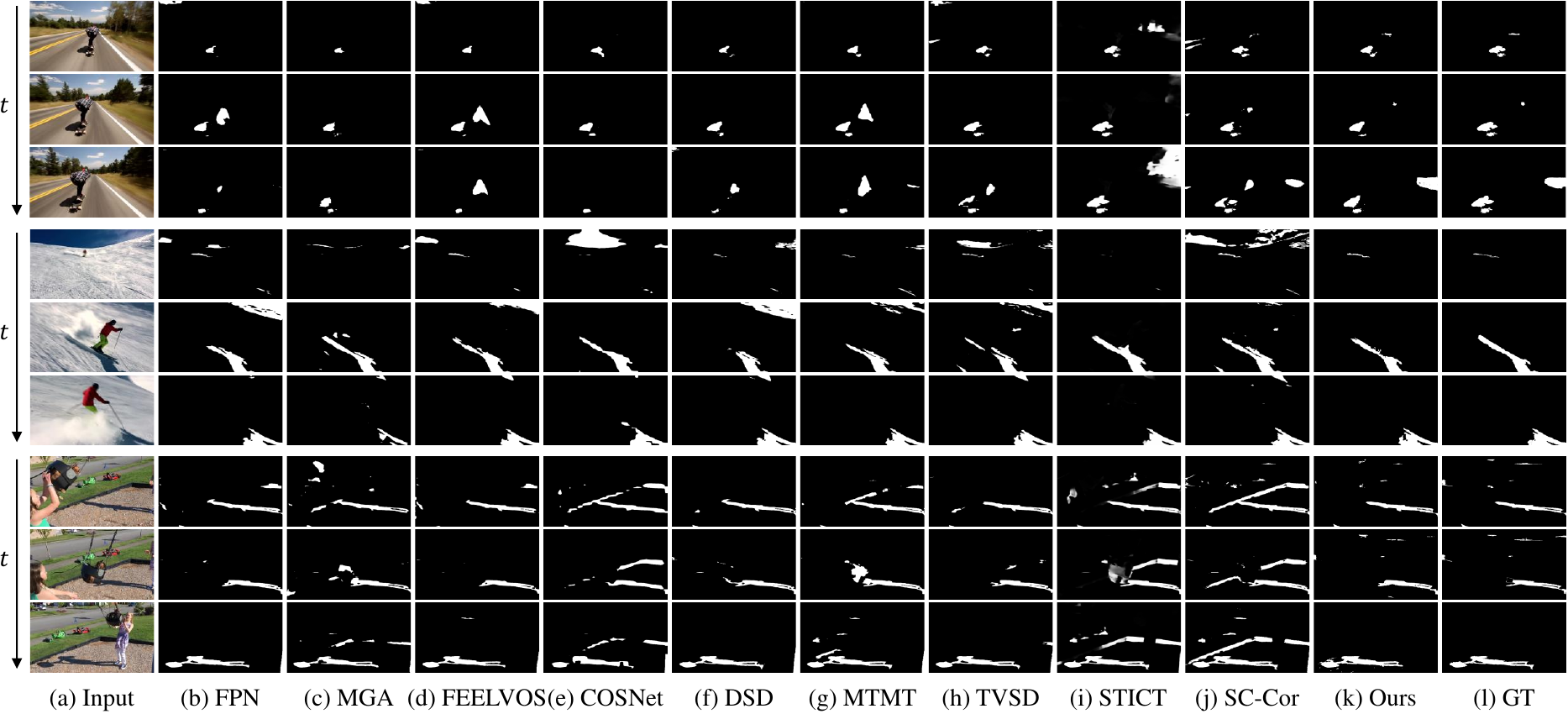}
	\caption{More visual comparison results of video shadow detection produced by different state-of-the-art methods. (b) to (j) are temporal prediction results from SOTA methods: FPN~\cite{lin2017feature}, MGA~\cite{li2019motion}, FEELVOS~\cite{voigtlaender2019feelvos}, COSNet~\cite{lu2019see}, DSD~\cite{zheng2019distraction}, MTMT~\cite{chen2020multi}, TVSD~\cite{chen2021triple}, STICT~\cite{lu2022video}, and SC-Cor~\cite{ding2022learning}, respectively.}
	\label{fig:more_results}
\end{figure*}

\begin{figure}[t]
	\centering
	\includegraphics[width=0.98\linewidth]{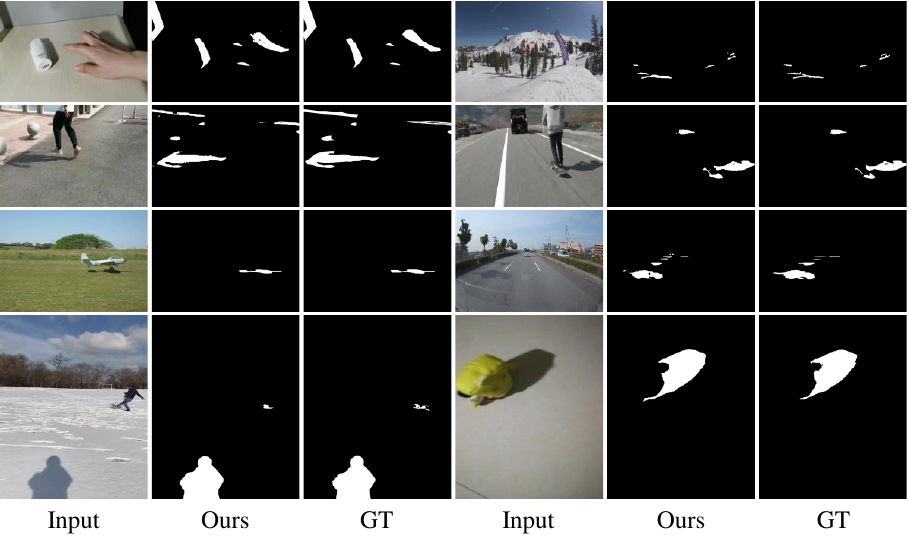}
	\caption{More visual results produced by our method.}
	\label{fig:pos}
\end{figure}

\noindent
\textbf{Qualitative comparison results.}
Figure~\ref{fig:qua1} illustrates the qualitative comparison results of our method with other state-of-the-art methods. 
For the first five rows, it can be observed that our method performs well in predicting ambiguous shadows, \emph{e.g.}, objects with similar colors to shadows or shadows that are not clearly visible, which other methods struggle to handle effectively.
Moreover, as indicated by the last three rows, our method exhibits favorable results in predicting small objects.
By leveraging the long short-term memory capability, the network can retain information about the number and approximate locations of shadows in the video, enabling accurate prediction.
Figure~\ref{fig:qua2} and Figure~\ref{fig:more_results} demonstrate the ability of the network to handle temporal information, revealing that our method continues to exhibit good performance in capturing temporal dependencies compared to other methods.
Furthermore, as shown in Figure~\ref{fig:pos}, we present additional visualization results generated by our method, further validating its reliability in various environmental scenes.

\setlength{\tabcolsep}{10pt}
\begin{table}[t]
	\caption{Comparison of model size (Params), computational complexity (GMACs), and inference time (Time) with the state-of-the-art VSD methods. ``*'' represents that the inference is performed on an A100 GPU, while our method is conducted on a 3090Ti GPU. The units for Params and Time are (MB) and (Mins), respectively.}
	\centering
	\resizebox{\linewidth}{!}{
		\begin{tabular}{l|ccc}
			\Xhline{2.5\arrayrulewidth}
			methods & Params (MB) & GMACs & Time (Mins)  \\
			\hline
			\hline
			TVSD*~\cite{chen2021triple}           & 243.32  & 158.89  & 32.4    \\
			STCIT*~\cite{lu2022video}             & 104.68  & 40.99   & 13.5    \\
			SC-Cor*~\cite{ding2022learning}       & 232.63  & 218.4   & 21.8    \\
			Liu \emph{et al.}*~\cite{liu2023scotch}         & 211.79  & 122.46   & 9.2  \\
			\hline
			Ours   						      &  93.73/7.60 &  427.57/9.41 &  8.8  \\    
			\Xhline{2.5\arrayrulewidth}
		\end{tabular}
	}
	\label{tab:efficiency}
\end{table}

\noindent
\textbf{Computational efficiency.}
Furthermore, we compared our network with the recent VSD method~\cite{chen2021triple,lu2022video,ding2022learning,liu2023scotch} in terms of model size (Params), computational complexity (GMACs), and inference time (Time).
The inference time represents the duration required to process the whole ViSha dataset~\cite{chen2021triple}, \emph{i.e.}, 70 videos with 6897 images.
As shown in Table~\ref{tab:efficiency}, the data is obtained from Liu \emph{et al.}~\cite{liu2023scotch}, where ``*" indicates that the inference time is calculated on an A100 GPU.
For comparison, our approach is calculated on an NVIDIA RTX 3090Ti GPU, which is slower.
In terms of model size, the fine-tuned SAM has 93.73M parameters, while the long short-term network is 7.60M.
Overall, our model has the fewest parameters among other VSD methods.
Regarding computational complexity, although SAM has a relatively high computational cost, our computational cost is very small during subsequent propagation (only 9.41 GMACs).
Additionally, the total inference time for our method on the whole test dataset is 8.8 minutes, which is the fastest among other video shadow detection methods, indicating the efficiency of our method.

\setlength{\tabcolsep}{10pt}
\begin{table}[t]
	\caption{Ablation study on the number of long short-term attention blocks. The best and the second results are highlighted in \textbf{bold} and \underline{underlined}, respectively. ``$\uparrow$'' indicates the higher the better and ``$\downarrow$'' means the opposite.}
	\centering
	\resizebox{\linewidth}{!}{
		\begin{tabular}{c|ccccc}
			\Xhline{2.5\arrayrulewidth}
			Blocks & MAE$\downarrow$ & $F_{\beta}$$\uparrow$ & IoU$\uparrow$  & BER$\downarrow$  & Time \\
			\hline
			\hline
			1    & 0.030  &  0.790  &  0.617  &  15.28  &  \textbf{1.7} \\
			2    & 0.026  &  0.793  &  0.620  &  14.65  &  \underline{2.5} \\    
			3    & \underline{0.021}  &  \underline{0.797}  &  \underline{0.640}  &  \underline{14.38}  &  3.3 \\
			4    & \textbf{0.020}  &  \textbf{0.799}  & \textbf{0.646}   &  \textbf{14.29}  &  4.2 \\    
			\Xhline{2.5\arrayrulewidth}
		\end{tabular}
	}
	\label{tab:abl_block}
\end{table}

\begin{table}[t]
	\caption{Ablation study on the component of long short-term attention. The best and the second results are highlighted in \textbf{bold} and \underline{underlined}, respectively. ``$\uparrow$'' indicates the higher the better and ``$\downarrow$'' means the opposite.}
	\centering
	\resizebox{\linewidth}{!}{
		\begin{tabular}{cc|cccc}
			\Xhline{2.5\arrayrulewidth}
			\multicolumn{2}{c|}{Configuration} & \multicolumn{4}{c}{Metrics}  \\
			\hline
			\hline
			Long & Short & MAE$\downarrow$ & $F_{\beta}$$\uparrow$ & IoU$\uparrow$  & BER$\downarrow$  \\
			\hline
			$\times$       & $\times$       &   0.056  &  0.657  &  0.480  &  19.98  \\
			$\times$       & $\checkmark$   &   \underline{0.026}  &  \underline{0.785}  &  \underline{0.619}  &  \underline{15.42}  \\
			$\checkmark$   & $\times$       &   0.027  &  0.756  &  0.600  &  15.88  \\    
			$\checkmark$   & $\checkmark$   &   \textbf{0.021}  &  \textbf{0.797}  &  \textbf{0.640}  &  \textbf{14.38}  \\    
			\Xhline{2.5\arrayrulewidth}
		\end{tabular}
	}
	\label{tab:abl_memory}
\end{table}

\setlength{\tabcolsep}{10pt}
\begin{table}[t]
	\caption{Validating the effectiveness of fine-tuned SAM. The best and the second results are highlighted in \textbf{bold} and \underline{underlined}, respectively. ``$\uparrow$'' indicates the higher the better and ``$\downarrow$'' means the opposite.}
	\centering
	\resizebox{\linewidth}{!}{
		\begin{tabular}{c|cccc}
			\Xhline{2.5\arrayrulewidth}
			Method & MAE$\downarrow$ & $F_{\beta}$$\uparrow$ & IoU$\uparrow$  & BER$\downarrow$  \\
			\hline
			\hline
			MTMT~\cite{chen2020multi}       & \underline{0.041}  &  \underline{0.748}  &  0.525  &  19.82  \\
			FSD~\cite{hu2021revisiting}     & 0.082  &  0.640  &  \underline{0.535}  &  \underline{15.32}  \\
			\hline
			Ours    & \textbf{0.021}  &  \textbf{0.797}  &  \textbf{0.640}  &  \textbf{14.38}  \\    
			\Xhline{2.5\arrayrulewidth}
		\end{tabular}
	}
	\label{tab:eff_sam}
\end{table}

\subsection{Ablation Study}
In this section, we conduct ablation experiments on the components of the long short-term network to verify their effectiveness.
The ablation experiments are divided into two parts: the number of long short-term attention blocks and the long short-term attention mechanism. 
In addition, we conduct another two experiments to verify the effectiveness of fine-tuned SAM and the generalization performance of our method.
All experiments are conducted by utilizing the mask of the first frame of the videos and propagate the video once.

\noindent
\textbf{Number of long short-term attention blocks.}
Table~\ref{tab:abl_block} demonstrates the impact of the number of long short-term attention blocks on performance. 
Our method outperforms TVSD~\cite{chen2021triple} in terms of both $F_{\beta}$ and IoU by only using one block while requiring only 1.7 minutes for inference on 70 videos comprising 6897 images (on a single 3090Ti GPU). 
As the number of blocks increases, our performance continues improving, and each additional block incurs a time overhead of less than 1 minute. 
In this paper, we choose block number 3 to balance the performance and speed.

\noindent
\textbf{Component of long short-term attention.}
As shown in Table~\ref{tab:abl_memory}, both long and short-term attention lead to a clear improvement in performance. 
Specifically, short-term attention contributes a little more than long-term.
Additionally, the combination of both yields the best performance, validating the effectiveness of the long short-term attention.

\begin{figure}[t]
	\centering
	\includegraphics[width=0.98\linewidth]{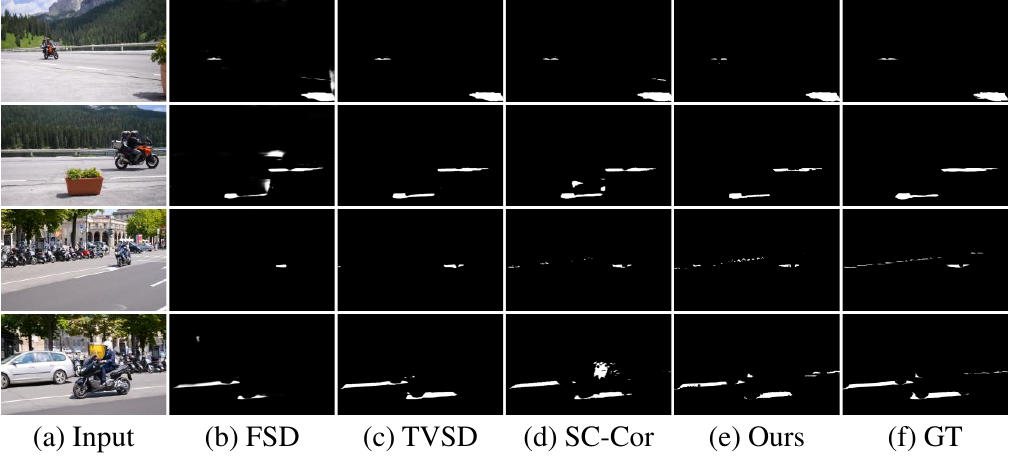}
	\caption{Visual comparison of our method with state-of-the-art video shadow detection method on the VISAD test dataset~\cite{lu2022video},  (b) to (d) are SOTA methods: TVSD~\cite{chen2021triple}, STICT~\cite{lu2022video}, and SC-Cor~\cite{ding2022learning}, respectively.}
	\label{fig:general}
\end{figure}

\setlength{\tabcolsep}{10pt}
\begin{table}[t]
	\caption{Quantitative comparison on the VISAD test dataset~\cite{lu2022video}. The best and the second results are highlighted in \textbf{bold} and \underline{underlined}, respectively. ``$\uparrow$'' indicates the higher the better and ``$\downarrow$'' means the opposite.}
	\centering
	\resizebox{\linewidth}{!}{
		\begin{tabular}{c|cccc}
			\Xhline{2.5\arrayrulewidth}
			Method & MAE$\downarrow$ & $F_{\beta}$$\uparrow$ & IoU$\uparrow$  & BER$\downarrow$  \\
			\hline
			\hline
			FSD~\cite{hu2021revisiting}     & 0.022  &  0.621  &  0.334  &  30.37  \\
			TVSD~\cite{chen2020multi}       & 0.021  &  0.654  &  0.341  &  31.87  \\
			SC-Cor~\cite{hu2021revisiting}     & \underline{0.018}  &  \underline{0.673}  &  \underline{0.426}  &  \underline{25.66}  \\
			\hline
			Ours    & \textbf{0.014}  & \textbf{0.723}   &  \textbf{0.507}  & \textbf{19.96}   \\    
			\Xhline{2.5\arrayrulewidth}
		\end{tabular}
	}
	\label{tab:gen_sam}
\end{table}

\begin{figure}[t]
	\centering
	\includegraphics[width=0.98\linewidth]{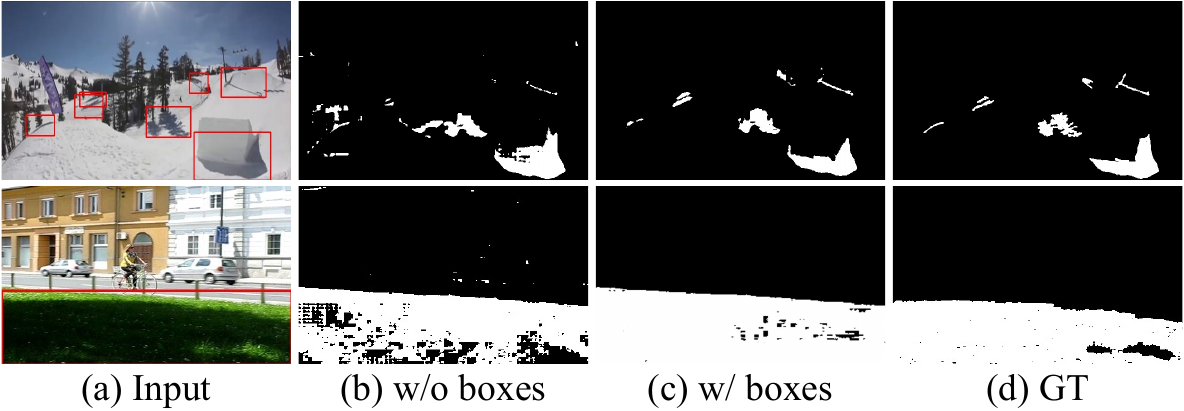}
	\caption{Visual comparison of our fine-tuned SAM w/o or w/ user interaction to provide bounding boxes of shadow region.}
	\label{fig:limit}
\end{figure}

\noindent
\textbf{The effectiveness of fine-tuned SAM.}
Our approach consists of two components: the fine-tuned SAM and the long short-term network.
Initially, the fine-tuned SAM is employed to predict the first frame of the video. 
Then, the long short-term network utilizes the mask of first frame as a shadow prior and propagates it to the subsequent frames. 
Specifically, the fine-tuned SAM can be viewed as a single-image shadow detector and can be replaced with other single-image detectors.
To verify the effectiveness of the fine-tuned SAM, we replace it with other single-frame shadow detection methods.
Here, we select two state-of-the-art single-image shadow detectors, namely MTMT~\cite{chen2020multi} and FSD~\cite{hu2021revisiting} to substitute the fine-tuned SAM.
MTMT~\cite{chen2020multi} employs a multi-task mean teacher model for semi-supervised shadow detection, which leverages additional unlabeled data and simultaneously learns multiple information about shadows.
FSD~\cite{hu2021revisiting} is a fast shadow detection network for detecting shadows in complex scenarios. 
We choose the two detectors due to their state-of-the-art performance and their capability to detect shadows in various complex situations.
In comparison, the fine-tuned SAM is a user-interaction network that is capable of detecting shadows using the provided bounding boxes of users.
We keep the rest of the ShadowSAM pipeline unchanged and set the number of long short-term attention blocks as 3.
The quantitative results are shown in Table~\ref{tab:eff_sam}, we can conclude that the performance of utilizing fine-tuned SAM outperforms that of other single-frame shadow detection methods, validating the effectiveness of fine-tuned SAM.

\noindent
\textbf{Generalization ability.}
To validate the generalization ability of our method, we randomly select 10 videos from the VISAD test dataset~\cite{lu2022video} and directly test our model on them.
We compare our method with state-of-the-art methods, including FSD~\cite{hu2021revisiting}, TVSD~\cite{chen2021triple}, and SC-Cor~\cite{ding2022learning}.
Table~\ref{tab:gen_sam} presents the quantitative comparison results, indicating that our method achieves the best performance among all four metrics.
In addition, Figure~\ref{fig:general} shows the visual comparison results, demonstrating that ShadowSAM exhibits better generalization ability in new environments.

%%%%%%%%%%%%%%%%%%%%%%%%%%%% discussion %%%%%%%%%%%%%%%%%%%%%%%%%%%%
\section{Discussion}
The proposed ShadowSAM achieves good performance in video shadow detection.
As shown in Figure~\ref{fig:pos}, we present more positive visual results produced by our method.
The results indicate that our method effectively detects shadows in various environments, such as roads, lawns, and snowy areas.

However, Our method has some limitations.
Firstly, as shown in Figure~\ref{fig:neg}, we provide some negative samples and the results indicate that our method may fail in the videos containing abundant and densely distributed small objects.
Considering the convenience in practical applications, providing bounding boxes for each small object in the first frame would impose a heavy burden on user interaction.
Therefore, when the number of shadows in the first frame exceeds eight, we only require the user to provide a bounding box encompassing all shadows (see Figure~\ref{fig:neg} row 1).
Such coarse bounding boxes may lead to imprecise shadow mask in the initial frame, and the subsequent error accumulation in the long short-term network could eventually result in the loss of some small object shadows.
Secondly, the fine-tuned SAM requires user interaction to provide bounding boxes in the first frame of the video.
As shown in Figure~\ref{fig:limit}, compared to the results without bounding boxes, incorporating bounding boxes allows SAM~\cite{kirillov2023segment} to predict shadows more accurately.
Notably, better detection of shadows in the first frame leads to more accurate video shadow detection results by the subsequent long short-term network. 
Given that for the task of video shadow detection, we only need to provide the bounding boxes of the first frame in the video, this is also acceptable.
Thirdly, even though we design the bounding boxes extraction method based on user interaction as much as possible, both the training and testing bounding boxes are still extracted from the ground truth shadow masks, which incorporate certain shadow prior information.

In future work, we aim to improve the fine-tuning strategy to enable SAM~\cite{kirillov2023segment} to detect shadows in images without providing bounding boxes, and we will also enhance the propagation strategy to achieve more accurate and efficient video shadow detection.

% \section{Conclusion and Future Work}
\section{Conclusion}
In this study, we present a new network architecture, called ShadowSAM, for video shadow detection. 
ShadowSAM is capable of predicting shadows in the first frame of a video and then propagates this shadow mask throughout the whole video.
Specifically, we design a simple yet effective approach to fine-tune SAM, enabling it to accurately detect shadows in natural images.
Subsequently, we employ a long short-term network to propagate the shadow mask from a single frame to the entire video.
We conduct extensive experiments on ViSha dataset, and the results demonstrate the effectiveness of our method, as well as a faster inference speed compared to other video shadow detection methods.

%
%\section*{Acknowledgments}
%This should be a simple paragraph before the References to thank those individuals and institutions who have supported your work on this article.

{\small
	\bibliographystyle{IEEEtran}
	\bibliography{egbib}

\begin{thebibliography}{10}
\providecommand{\url}[1]{#1}
\csname url@samestyle\endcsname
\providecommand{\newblock}{\relax}
\providecommand{\bibinfo}[2]{#2}
\providecommand{\BIBentrySTDinterwordspacing}{\spaceskip=0pt\relax}
\providecommand{\BIBentryALTinterwordstretchfactor}{4}
\providecommand{\BIBentryALTinterwordspacing}{\spaceskip=\fontdimen2\font plus
\BIBentryALTinterwordstretchfactor\fontdimen3\font minus
  \fontdimen4\font\relax}
\providecommand{\BIBforeignlanguage}[2]{{%
\expandafter\ifx\csname l@#1\endcsname\relax
\typeout{** WARNING: IEEEtran.bst: No hyphenation pattern has been}%
\typeout{** loaded for the language `#1'. Using the pattern for}%
\typeout{** the default language instead.}%
\else
\language=\csname l@#1\endcsname
\fi
#2}}
\providecommand{\BIBdecl}{\relax}
\BIBdecl

\bibitem{lalonde2009est}
J.-F. Lalonde, A.~A. Efros, and S.~G. Narasimhan, ``Estimating natural
  illumination from a single outdoor image,'' in \emph{Proceedings of the IEEE
  International Conference on Computer Vision}, 2009, pp. 183--190.

\bibitem{lalonde2012estimating}
{Lalonde, Jean-Fran{\c{c}}ois and Efros, Alexei A and Narasimhan, Srinivasa G},
  ``Estimating the natural illumination conditions from a single outdoor
  image,'' \emph{International Journal of Computer Vision}, vol.~98, pp.
  123--145, 2012.

\bibitem{panagopoulos2009robust}
A.~Panagopoulos, D.~Samaras, and N.~Paragios, ``Robust shadow and illumination
  estimation using a mixture model,'' in \emph{Proceedings of the IEEE
  Conference on Computer Vision and Pattern Recognition}, 2009, pp. 651--658.

\bibitem{huang2011what}
X.~Huang, G.~Hua, J.~Tumblin, and L.~Williams, ``What characterizes a shadow
  boundary under the sun and sky?'' in \emph{Proceedings of the IEEE
  International Conference on Computer Vision}, 2011, pp. 898--905.

\bibitem{takahiro2009attached}
T.~Okabe, I.~Sato, and Y.~Sato, ``Attached shadow coding: Estimating surface
  normals from shadows under unknown reflectance and lighting conditions,'' in
  \emph{Proceedings of the IEEE International Conference on Computer Vision},
  2009, pp. 1693--1700.

\bibitem{junejo2008estimating}
I.~N. Junejo and H.~Foroosh, ``Estimating geo-temporal location of stationary
  cameras using shadow trajectories,'' in \emph{Proceedings of the European
  Conference on Computer Vision}, 2008, pp. 318--331.

\bibitem{okabe2009attached}
T.~Okabe, I.~Sato, and Y.~Sato, ``Attached shadow coding: Estimating surface
  normals from shadows under unknown reflectance and lighting conditions,'' in
  \emph{Proceedings of the IEEE International Conference on Computer Vision},
  2009, pp. 1693--1700.

\bibitem{thien2003image}
C.-C. Thien and J.-C. Lin, ``An image-sharing method with user-friendly shadow
  images,'' \emph{IEEE Transactions on Circuits and Systems for Video
  Technology}, vol.~13, no.~12, pp. 1161--1169, 2003.

\bibitem{wu2010camera}
L.~Wu, X.~Cao, and H.~Foroosh, ``Camera calibration and geo-location estimation
  from two shadow trajectories,'' \emph{Computer Vision and Image
  Understanding}, vol. 114, no.~8, pp. 915--927, 2010.

\bibitem{cucchiara2003detecting}
R.~Cucchiara, C.~Grana, M.~Piccardi, and A.~Prati, ``Detecting moving objects,
  ghosts, and shadows in video streams,'' \emph{IEEE Transactions on Pattern
  Analysis and Machine Intelligence}, vol.~25, no.~10, pp. 1337--1342, 2003.

\bibitem{chien2002efficient}
S.-Y. Chien, S.-Y. Ma, and L.-G. Chen, ``Efficient moving object segmentation
  algorithm using background registration technique,'' \emph{IEEE Transactions
  on Circuits and Systems for Video Technology}, vol.~12, no.~7, pp. 577--586,
  2002.

\bibitem{ecins2014shadow}
A.~Ecins, C.~Ferm{\"u}ller, and Y.~Aloimonos, ``Shadow free segmentation in
  still images using local density measure,'' in \emph{Proceedings of the IEEE
  International Conference on Computational Photography}, 2014, pp. 1--8.

\bibitem{liu2020arshadowgan}
D.~Liu, C.~Long, H.~Zhang, H.~Yu, X.~Dong, and C.~Xiao, ``{ARShadowGAN}: Shadow
  generative adversarial network for augmented reality in single light
  scenes,'' in \emph{Proceedings of the IEEE Conference on Computer Vision and
  Pattern Recognition}, 2020, pp. 8139--8148.

\bibitem{vicente2016large}
T.~F.~Y. Vicente, L.~Hou, C.-P. Yu, M.~Hoai, and D.~Samaras, ``Large-scale
  training of shadow detectors with noisily-annotated shadow examples,'' in
  \emph{Proceedings of the European Conference on Computer Vision}, 2016, pp.
  816--832.

\bibitem{hu2018direction}
X.~Hu, L.~Zhu, C.-W. Fu, J.~Qin, and P.-A. Heng, ``Direction-aware spatial
  context features for shadow detection,'' in \emph{Proceedings of the IEEE
  Conference on Computer Vision and Pattern Recognition}, 2018, pp. 7454--7462.

\bibitem{zhu2018bidirectional}
L.~Zhu, Z.~Deng, X.~Hu, C.-W. Fu, X.~Xu, J.~Qin, and P.-A. Heng,
  ``Bidirectional feature pyramid network with recurrent attention residual
  modules for shadow detection,'' in \emph{Proceedings of the European
  Conference on Computer Vision}, 2018, pp. 121--136.

\bibitem{zheng2019distraction}
Q.~Zheng, X.~Qiao, Y.~Cao, and R.~W. Lau, ``Distraction-aware shadow
  detection,'' in \emph{Proceedings of the IEEE Conference on Computer Vision
  and Pattern Recognition}, 2019, pp. 5167--5176.

\bibitem{chen2020multi}
Z.~Chen, L.~Zhu, L.~Wan, S.~Wang, W.~Feng, and P.-A. Heng, ``A multi-task mean
  teacher for semi-supervised shadow detection,'' in \emph{Proceedings of the
  IEEE Conference on Computer Vision and Pattern Recognition}, 2020, pp.
  5611--5620.

\bibitem{zhu2021mitigating}
L.~Zhu, K.~Xu, Z.~Ke, and R.~W. Lau, ``Mitigating intensity bias in shadow
  detection via feature decomposition and reweighting,'' in \emph{Proceedings
  of the IEEE Conference on Computer Vision and Pattern Recognition}, 2021, pp.
  4702--4711.

\bibitem{chen2021triple}
Z.~Chen, L.~Wan, L.~Zhu, J.~Shen, H.~Fu, W.~Liu, and J.~Qin,
  ``Triple-cooperative video shadow detection,'' in \emph{Proceedings of the
  IEEE Conference on Computer Vision and Pattern Recognition}, 2021, pp.
  2715--2724.

\bibitem{ding2022learning}
X.~Ding, J.~Yang, X.~Hu, and X.~Li, ``Learning shadow correspondence for video
  shadow detection,'' in \emph{Proceedings of the European Conference on
  Computer Vision}, 2022, pp. 705--722.

\bibitem{pont20172017}
J.~Pont-Tuset, F.~Perazzi, S.~Caelles, P.~Arbel{\'a}ez, A.~Sorkine-Hornung, and
  L.~Van~Gool, ``The 2017 davis challenge on video object segmentation,''
  \emph{arXiv preprint arXiv:1704.00675}, 2017.

\bibitem{xu2018youtube}
N.~Xu, L.~Yang, Y.~Fan, D.~Yue, Y.~Liang, J.~Yang, and T.~Huang,
  ``{YouTube-VOS}: A large-scale video object segmentation benchmark,''
  \emph{arXiv preprint arXiv:1809.03327}, 2018.

\bibitem{yang2021associating}
Z.~Yang, Y.~Wei, and Y.~Yang, ``Associating objects with transformers for video
  object segmentation,'' \emph{Advances in Neural Information Processing
  Systems}, vol.~34, pp. 2491--2502, 2021.

\bibitem{cheng2022xmem}
H.~K. Cheng and A.~G. Schwing, ``{XMem}: Long-term video object segmentation
  with an atkinson-shiffrin memory model,'' in \emph{Proceedings of the
  European Conference on Computer Vision}, 2022, pp. 640--658.

\bibitem{xu2022reliable}
X.~Xu, J.~Wang, X.~Li, and Y.~Lu, ``Reliable propagation-correction modulation
  for video object segmentation,'' in \emph{Proceedings of the AAAI Conference
  on Artificial Intelligence}, 2022, pp. 2946--2954.

\bibitem{kirillov2023segment}
A.~Kirillov, E.~Mintun, N.~Ravi, H.~Mao, C.~Rolland, L.~Gustafson, T.~Xiao,
  S.~Whitehead, A.~C. Berg, W.-Y. Lo \emph{et~al.}, ``Segment anything,''
  \emph{arXiv preprint arXiv:2304.02643}, 2023.

\bibitem{gao2022aiatrack}
S.~Gao, C.~Zhou, C.~Ma, X.~Wang, and J.~Yuan, ``{AiATrack}: Attention in
  attention for transformer visual tracking,'' in \emph{Proceedings of the
  European Conference on Computer Vision}, 2022, pp. 146--164.

\bibitem{mao2021joint}
Y.~Mao, N.~Wang, W.~Zhou, and H.~Li, ``Joint inductive and transductive
  learning for video object segmentation,'' in \emph{Proceedings of the IEEE
  International Conference on Computer Vision}, 2021, pp. 9670--9679.

\bibitem{liu2023scotch}
L.~Liu, J.~Prost, L.~Zhu, N.~Papadakis, P.~Li{\`o}, C.-B. Sch{\"o}nlieb, and
  A.~I. Aviles-Rivero, ``{SCOTCH and SODA}: A transformer video shadow
  detection framework,'' in \emph{Proceedings of the IEEE Conference on
  Computer Vision and Pattern Recognition}, 2023, pp. 10\,449--10\,458.

\bibitem{hu2021revisiting}
X.~Hu, T.~Wang, C.-W. Fu, Y.~Jiang, Q.~Wang, and P.-A. Heng, ``Revisiting
  shadow detection: A new benchmark dataset for complex world,'' \emph{IEEE
  Transactions on Image Processing}, vol.~30, pp. 1925--1934, 2021.

\bibitem{inoue2020learning}
N.~Inoue and T.~Yamasaki, ``Learning from synthetic shadows for shadow
  detection and removal,'' \emph{IEEE Transactions on Circuits and Systems for
  Video Technology}, vol.~31, no.~11, pp. 4187--4197, 2020.

\bibitem{jie2023rmlanet}
L.~Jie and H.~Zhang, ``{RMLANet}: Random multi-level attention network for
  shadow detection and removal,'' \emph{IEEE Transactions on Circuits and
  Systems for Video Technology}, 2023.

\bibitem{nadimi2004physical}
S.~Nadimi and B.~Bhanu, ``Physical models for moving shadow and object
  detection in video,'' \emph{IEEE Transactions on Pattern Analysis and Machine
  Intelligence}, vol.~26, no.~8, pp. 1079--1087, 2004.

\bibitem{stander1999detection}
J.~Stander, R.~Mech, and J.~Ostermann, ``Detection of moving cast shadows for
  object segmentation,'' \emph{IEEE Transactions on Multimedia}, vol.~1, no.~1,
  pp. 65--76, 1999.

\bibitem{elgammal2002background}
A.~Elgammal, R.~Duraiswami, D.~Harwood, and L.~Davis, ``Background and
  foreground modeling using nonparametric kernel density estimation for visual
  surveillance,'' \emph{Proceedings of the IEEE}, vol.~90, no.~7, pp.
  1151--1163, 2002.

\bibitem{jacques2005background}
J.~C.~S. Jacques, C.~R. Jung, and S.~R. Musse, ``Background subtraction and
  shadow detection in grayscale video sequences,'' in \emph{Proceedings of the
  Conference on Graphics, Patterns and Images}, 2005, pp. 189--196.

\bibitem{kumar2002comparative}
P.~Kumar, K.~Sengupta, and A.~Lee, ``A comparative study of different color
  spaces for foreground and shadow detection for traffic monitoring system,''
  in \emph{Proceedings of the IEEE Conference on Intelligent Transportation
  Systems}, 2002, pp. 100--105.

\bibitem{salvador2004cast}
E.~Salvador, A.~Cavallaro, and T.~Ebrahimi, ``Cast shadow segmentation using
  invariant color features,'' \emph{Computer Vision and Image Understanding},
  vol.~95, no.~2, pp. 238--259, 2004.

\bibitem{xu2005insignificant}
D.~Xu, J.~Liu, X.~Li, Z.~Liu, and X.~Tang, ``Insignificant shadow detection for
  video segmentation,'' \emph{IEEE Transactions on Circuits and Systems for
  Video Technology}, vol.~15, no.~8, pp. 1058--1064, 2005.

\bibitem{wang2009real}
Y.~Wang, ``Real-time moving vehicle detection with cast shadow removal in video
  based on conditional random field,'' \emph{IEEE Transactions on Circuits and
  Systems for Video Technology}, vol.~19, no.~3, pp. 437--441, 2009.

\bibitem{liu2011cast}
Z.~Liu, K.~Huang, and T.~Tan, ``Cast shadow removal in a hierarchical manner
  using mrf,'' \emph{IEEE Transactions on Circuits and Systems for Video
  Technology}, vol.~22, no.~1, pp. 56--66, 2011.

\bibitem{russell2017feature}
M.~Russell, J.~J. Zou, G.~Fang, and W.~Cai, ``Feature-based image patch
  classification for moving shadow detection,'' \emph{IEEE Transactions on
  Circuits and Systems for Video Technology}, vol.~29, no.~9, pp. 2652--2666,
  2017.

\bibitem{lu2022video}
X.~Lu, Y.~Cao, S.~Liu, C.~Long, Z.~Chen, X.~Zhou, Y.~Yang, and C.~Xiao, ``Video
  shadow detection via spatio-temporal interpolation consistency training,'' in
  \emph{Proceedings of the IEEE Conference on Computer Vision and Pattern
  Recognition}, 2022, pp. 3116--3125.

\bibitem{vaswani2017attention}
A.~Vaswani, N.~Shazeer, N.~Parmar, J.~Uszkoreit, L.~Jones, A.~N. Gomez,
  {\L}.~Kaiser, and I.~Polosukhin, ``Attention is all you need,''
  \emph{Advances in Neural Information Processing Systems}, vol.~30, 2017.

\bibitem{dosovitskiyimage}
A.~Dosovitskiy, L.~Beyer, A.~Kolesnikov, D.~Weissenborn, X.~Zhai,
  T.~Unterthiner, M.~Dehghani, M.~Minderer, G.~Heigold, S.~Gelly \emph{et~al.},
  ``An image is worth 16x16 words: Transformers for image recognition at
  scale,'' in \emph{Proceedings of the IEEE Conference on International
  Conference on Learning Representations}, 2021.

\bibitem{carion2020end}
N.~Carion, F.~Massa, G.~Synnaeve, N.~Usunier, A.~Kirillov, and S.~Zagoruyko,
  ``End-to-end object detection with transformers,'' in \emph{Proceedings of
  the European Conference on Computer Vision}, 2020, pp. 213--229.

\bibitem{wang2021end}
Y.~Wang, Z.~Xu, X.~Wang, C.~Shen, B.~Cheng, H.~Shen, and H.~Xia, ``End-to-end
  video instance segmentation with transformers,'' in \emph{Proceedings of the
  IEEE Conference on Computer Vision and Pattern Recognition}, 2021, pp.
  8741--8750.

\bibitem{zhu2021deformable}
X.~Zhu, W.~Su, L.~Lu, B.~Li, X.~Wang, and J.~Dai, ``{Deformable DETR}:
  Deformable transformers for end-to-end object detection,'' in
  \emph{Proceedings of the IEEE Conference on International Conference on
  Learning Representations}, 2021.

\bibitem{liu2021swin}
Z.~Liu, Y.~Lin, Y.~Cao, H.~Hu, Y.~Wei, Z.~Zhang, S.~Lin, and B.~Guo, ``Swin
  transformer: Hierarchical vision transformer using shifted windows,'' in
  \emph{Proceedings of the IEEE International Conference on Computer Vision},
  2021, pp. 10\,012--10\,022.

\bibitem{bommasani2021opportunities}
R.~Bommasani, D.~A. Hudson, E.~Adeli, R.~Altman, S.~Arora, S.~von Arx, M.~S.
  Bernstein, J.~Bohg, A.~Bosselut, E.~Brunskill \emph{et~al.}, ``On the
  opportunities and risks of foundation models,'' \emph{arXiv preprint
  arXiv:2108.07258}, 2021.

\bibitem{devlin2019bert}
J.~Devlin, M.-W. Chang, K.~Lee, and K.~Toutanova, ``{BERT}: Pre-training of
  deep bidirectional transformers for language understanding,'' in
  \emph{Association for Computational Linguistics}, 2019, pp. 4171--4186.

\bibitem{radford2019language}
A.~Radford, J.~Wu, R.~Child, D.~Luan, D.~Amodei, I.~Sutskever \emph{et~al.},
  ``Language models are unsupervised multitask learners,'' \emph{OpenAI blog},
  vol.~1, no.~8, p.~9, 2019.

\bibitem{liu2019roberta}
Y.~Liu, M.~Ott, N.~Goyal, J.~Du, M.~Joshi, D.~Chen, O.~Levy, M.~Lewis,
  L.~Zettlemoyer, and V.~Stoyanov, ``{RoBERTa}: A robustly optimized bert
  pretraining approach,'' \emph{arXiv preprint arXiv:1907.11692}, 2019.

\bibitem{raffel2020exploring}
C.~Raffel, N.~Shazeer, A.~Roberts, K.~Lee, S.~Narang, M.~Matena, Y.~Zhou,
  W.~Li, and P.~J. Liu, ``Exploring the limits of transfer learning with a
  unified text-to-text transformer,'' \emph{Journal of Machine Learning
  Research}, vol.~21, no.~1, pp. 5485--5551, 2020.

\bibitem{radford2021learning}
A.~Radford, J.~W. Kim, C.~Hallacy, A.~Ramesh, G.~Goh, S.~Agarwal, G.~Sastry,
  A.~Askell, P.~Mishkin, J.~Clark \emph{et~al.}, ``Learning transferable visual
  models from natural language supervision,'' in \emph{Proceedings of the IEEE
  Conference on International Conference on Machine Learning}, 2021, pp.
  8748--8763.

\bibitem{wang2022image}
W.~Wang, H.~Bao, L.~Dong, J.~Bjorck, Z.~Peng, Q.~Liu, K.~Aggarwal, O.~K.
  Mohammed, S.~Singhal, S.~Som \emph{et~al.}, ``Image as a foreign language:
  Beit pretraining for all vision and vision-language tasks,'' \emph{arXiv
  preprint arXiv:2208.10442}, 2022.

\bibitem{he2022masked}
K.~He, X.~Chen, S.~Xie, Y.~Li, P.~Doll{\'a}r, and R.~Girshick, ``Masked
  autoencoders are scalable vision learners,'' in \emph{Proceedings of the IEEE
  Conference on Computer Vision and Pattern Recognition}, 2022, pp.
  16\,000--16\,009.

\bibitem{sandler2018mobilenetv2}
M.~Sandler, A.~Howard, M.~Zhu, A.~Zhmoginov, and L.-C. Chen, ``{MobileNetV2}:
  Inverted residuals and linear bottlenecks,'' in \emph{Proceedings of the IEEE
  Conference on Computer Vision and Pattern Recognition}, 2018, pp. 4510--4520.

\bibitem{lin2017feature}
T.-Y. Lin, P.~Doll{\'a}r, R.~Girshick, K.~He, B.~Hariharan, and S.~Belongie,
  ``Feature pyramid networks for object detection,'' in \emph{Proceedings of
  the IEEE Conference on Computer Vision and Pattern Recognition}, 2017, pp.
  2117--2125.

\bibitem{hendrycks2016gaussian}
D.~Hendrycks and K.~Gimpel, ``Gaussian error linear units ({GELUs}),''
  \emph{arXiv preprint arXiv:1606.08415}, 2016.

\bibitem{zhao2017pyramid}
H.~Zhao, J.~Shi, X.~Qi, X.~Wang, and J.~Jia, ``Pyramid scene parsing network,''
  in \emph{Proceedings of the IEEE Conference on Computer Vision and Pattern
  Recognition}, 2017, pp. 2881--2890.

\bibitem{hou2017deeply}
Q.~Hou, M.-M. Cheng, X.~Hu, A.~Borji, Z.~Tu, and P.~H. Torr, ``Deeply
  supervised salient object detection with short connections,'' in
  \emph{Proceedings of the IEEE Conference on Computer Vision and Pattern
  Recognition}, 2017, pp. 3203--3212.

\bibitem{li2019motion}
H.~Li, G.~Chen, G.~Li, and Y.~Yu, ``Motion guided attention for video salient
  object detection,'' in \emph{Proceedings of the IEEE International Conference
  on Computer Vision}, 2019, pp. 7274--7283.

\bibitem{song2018pyramid}
H.~Song, W.~Wang, S.~Zhao, J.~Shen, and K.-M. Lam, ``Pyramid dilated deeper
  convlstm for video salient object detection,'' in \emph{Proceedings of the
  European Conference on Computer Vision}, 2018, pp. 715--731.

\bibitem{voigtlaender2019feelvos}
P.~Voigtlaender, Y.~Chai, F.~Schroff, H.~Adam, B.~Leibe, and L.-C. Chen,
  ``{FEELVOS}: Fast end-to-end embedding learning for video object
  segmentation,'' in \emph{Proceedings of the IEEE Conference on Computer
  Vision and Pattern Recognition}, 2019, pp. 9481--9490.

\bibitem{oh2019video}
S.~W. Oh, J.-Y. Lee, N.~Xu, and S.~J. Kim, ``Video object segmentation using
  space-time memory networks,'' in \emph{Proceedings of the IEEE International
  Conference on Computer Vision}, 2019, pp. 9226--9235.

\bibitem{lu2019see}
X.~Lu, W.~Wang, C.~Ma, J.~Shen, L.~Shao, and F.~Porikli, ``See more, know more:
  Unsupervised video object segmentation with co-attention siamese networks,''
  in \emph{Proceedings of the IEEE Conference on Computer Vision and Pattern
  Recognition}, 2019, pp. 3623--3632.

\bibitem{ba2016layer}
J.~L. Ba, J.~R. Kiros, and G.~E. Hinton, ``Layer normalization,'' \emph{arXiv
  preprint arXiv:1607.06450}, 2016.

\bibitem{nowozin2014optimal}
S.~Nowozin, ``Optimal decisions from probabilistic models: the
  intersection-over-union case,'' in \emph{Proceedings of the IEEE Conference
  on Computer Vision and Pattern Recognition}, 2014, pp. 548--555.

\bibitem{rahman2016optimizing}
M.~A. Rahman and Y.~Wang, ``Optimizing intersection-over-union in deep neural
  networks for image segmentation,'' in \emph{Proceedings of the IEEE
  Conference on Intelligent Systems and Computer Vision}, 2016, pp. 234--244.

\bibitem{deng2009imagenet}
J.~Deng, W.~Dong, R.~Socher, L.-J. Li, K.~Li, and L.~Fei-Fei, ``Imagenet: A
  large-scale hierarchical image database,'' in \emph{Proceedings of the IEEE
  Conference on Computer Vision and Pattern Recognition}, 2009, pp. 248--255.

\bibitem{kingma2014adam}
D.~P. Kingma and J.~Ba, ``Adam: A method for stochastic optimization,''
  \emph{arXiv preprint arXiv:1412.6980}, 2014.

\bibitem{patrick2021keeping}
M.~Patrick, D.~Campbell, Y.~Asano, I.~Misra, F.~Metze, C.~Feichtenhofer,
  A.~Vedaldi, and J.~F. Henriques, ``Keeping your eye on the ball: Trajectory
  attention in video transformers,'' \emph{Advances in Neural Information
  Processing Systems}, vol.~34, pp. 12\,493--12\,506, 2021.

\end{thebibliography}
}

\vfill

\end{document}